\documentclass[10pt,twocolumn,letterpaper]{article}

\usepackage{iccv}
\usepackage{times}
\usepackage{epsfig}
\usepackage{graphicx}
\usepackage{amsmath}
\usepackage{amssymb}
\newcommand\norm[1]{\left\lVert#1\right\rVert}
\usepackage[caption=false]{subfig}
\usepackage{threeparttable}
\usepackage[ruled]{algorithm2e}
\usepackage{array}
\usepackage{hhline}
\usepackage{xcolor,colortbl}
\definecolor{Gray}{gray}{0.94}

\usepackage[pagebackref=true,breaklinks=true,letterpaper=true,colorlinks,bookmarks=false]{hyperref}

\iccvfinalcopy 

\def\iccvPaperID{5088} 

\ificcvfinal\fi
\begin{document}

\title{Distilling Knowledge From a Deep Pose Regressor Network}

\author{Muhamad Risqi U. Saputra,  Pedro P. B. de Gusmao, Yasin Almalioglu, Andrew Markham, Niki Trigoni\\
Department of Computer Science, University of Oxford\\
{\tt\small firstname.lastname@cs.ox.ac.uk}
}

\maketitle

\begin{abstract}
This paper presents a novel method to distill knowledge from a deep pose regressor network for efficient Visual Odometry (VO). Standard distillation relies on ``dark knowledge'' for successful knowledge transfer. As this knowledge is not available in pose regression and the teacher prediction is not always accurate, we propose to emphasize the knowledge transfer only when we trust the teacher. We achieve this by using teacher loss as a confidence score which places variable relative importance on the teacher prediction. We inject this confidence score to the main training task via Attentive Imitation Loss (AIL) and when learning the intermediate representation of the teacher through Attentive Hint Training (AHT) approach. To the best of our knowledge, this is the first work which successfully distill the knowledge from a deep pose regression network. Our evaluation on the KITTI and Malaga dataset shows that we can keep the student prediction close to the teacher with up to $92.95\%$ parameter reduction and $2.12\times$ faster in computation time.
\end{abstract}

\section{Introduction}

Deep Neural Networks (DNNs) have received increased attention in the last decade due to their success in image and natural language understanding. The availability of large datasets, increased computing power, and advancement of learning algorithms play a pivotal role in their glory. Despite their successes, DNN-based approaches typically require tens or hundreds of million weights. As a consequence, the huge computational and space requirement prevents DNN models from being widely implemented in resource-constrained environment (e.g. mobile phones, quadcopter, etc.). To compound the issue, these applications typically require near real-time inference.

Within the last few years, there have been tremendous efforts towards compressing DNNs. State-of-the-art approaches for network compression such as quantization \cite{Gong2014, Courbariaux2015, Hubara2016a}, pruning \cite{Han2015b, Guo2016, Yao2017}, or low-rank decomposition \cite{Tai2015, Bhattacharya2016} can yield significant speed-ups but at the cost of accuracy. On the other hand, an approach called Knowledge Distillation (KD) proposed by Hinton et al. \cite{Hinton2015} offers to recover the accuracy drop by transferring the knowledge of a large teacher model to a small student model. Some recent works show that a small network trained by KD could match or even exceed the accuracy of a large network if it is trained with careful optimization \cite{Romero2014}.

Most works in network compression, including KD, focus on the problem of classification. KD works very well in classification since it has the advantage of ``dark knowledge" which refers to the softened logits output of the teacher. This provides more information than mere one-hot encoding of the class label and contains hidden knowledge about the correlations of class labels \cite{Hinton2015}. By using the logits output for training, the student network can emulate the generalization capability of the teacher network. However, this advantage does not exist in \textbf{regression}. In the regression problem, a deep regression network predicts sequential, continuous, values which have the exact same characteristics as the ground truth, with the exception of being plagued with an unknown error distribution. Without access to any dark knowledge, it is unclear how KD could help in compressing a regression network. In recent surveys, it is even stated that the main drawback of KD is that it only works for classification problems \cite{cheng2018}. 

\begin{figure*}[!t]
    \centering
    \subfloat[KD in Classification]{
        \begin{tabular}[b]{c}%
        \includegraphics[width=0.8\columnwidth]{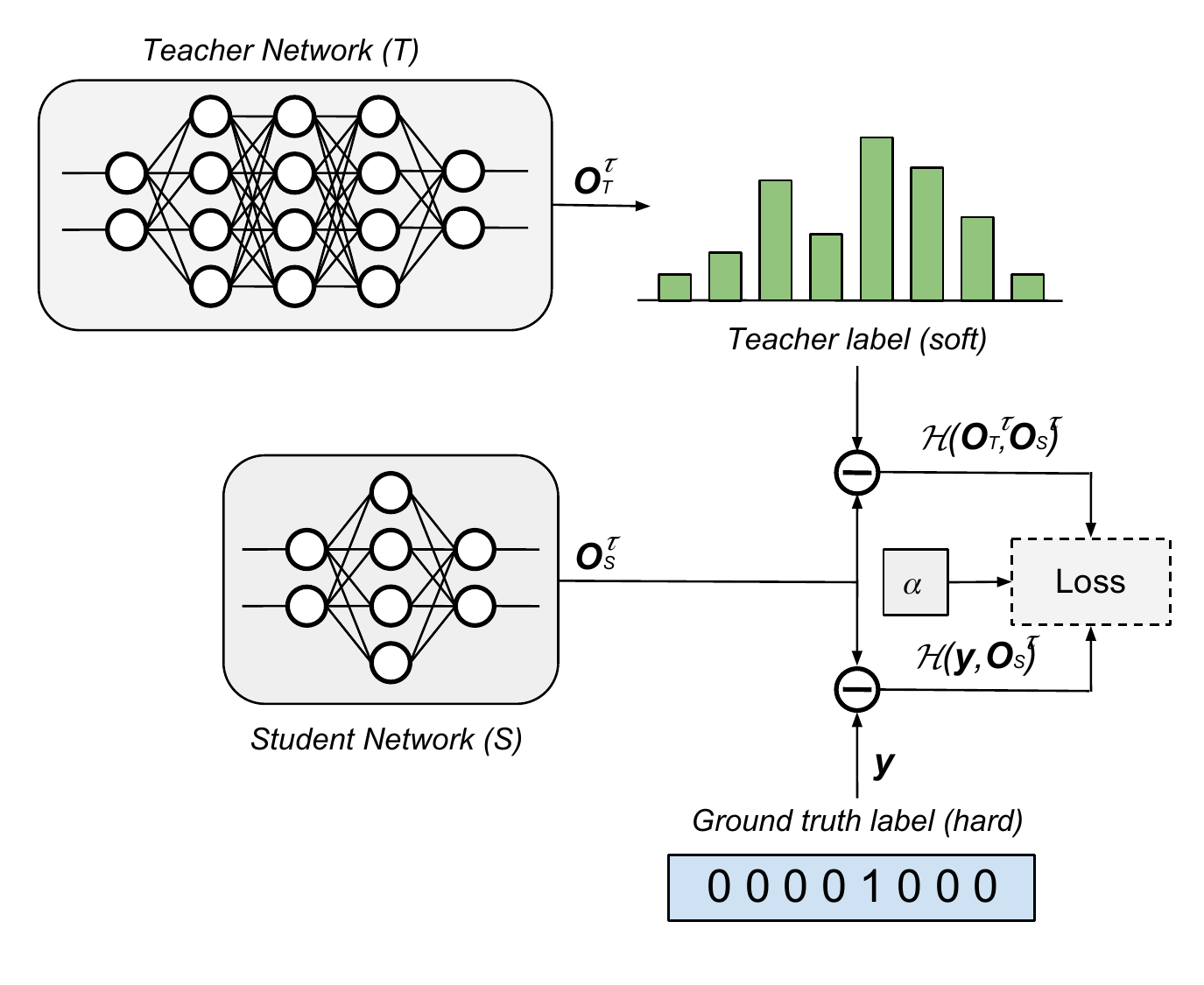}
        \end{tabular}
        }
    \subfloat[KD in Regression]{
        \begin{tabular}[b]{c}%
    	\includegraphics[width=1.1\columnwidth]{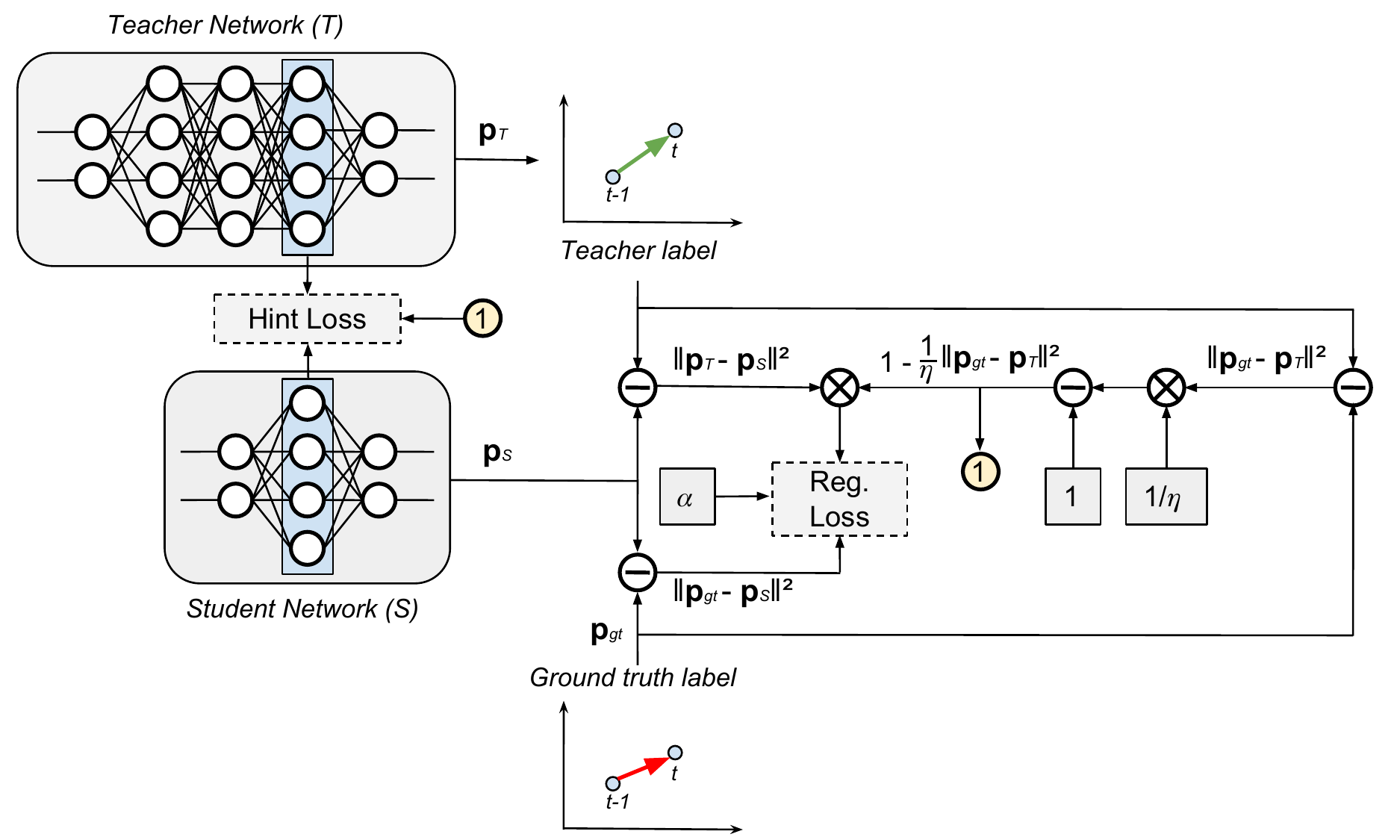}
        \end{tabular}
        }
    \caption{Comparison between (a) standard Knowledge Distillation applied to classification problem and (b) our Knowledge Distillation approach applied to regression problem. Note that in regression, we are unable to use the dark knowledge provided by soft teacher labels.}
\label{fig:different_clas_reg}
\end{figure*}

KD methods for classification \cite{Hinton2015, Romero2014, lopez2016unifying, yim2017gift, wang2018progressive, lee2018self, polino2018model} rely solely on the teacher prediction without considering the error made w.r.t. ground truth. In regression however, the real-valued predictions are unbounded, and hence the teacher can give highly erroneous guidance to the student network. Previous work \cite{chen2017learning} alleviated this issue by using teacher loss as an upper bound. However, it was designed for standard bounding box regression which has different characteristic to pose regression as it belongs to $SE(3)$ (Lie Groups). Moreover, they directly transferred the knowledge from the teacher without filtering which one is good and which one is bad. To this end, our novel insight is to use the teacher loss as a confidence score to decide when we can trust the teacher. We demonstrate that this is key to successfully distilling deep pose regression networks.

We will demonstrate our work in distillation for camera pose regression problem which is widely known as Visual Odometry (VO). In particular, we employ DNN-based VO methods \cite{wang2017, Wang2017a, zhan2018unsupervised, zhou2018deeptam, saputra19clvo, almalioglu19} which replaces the conventional VO pipeline based on multiple-view geometry \cite{Hartley2004, Saputra2018} to DNN-based pipeline which automatically learns useful features for estimating 6 Degree-of-Freedom (DoF) camera poses. To the best of our knowledge, this work is a first attempt to distill the knowledge from a deep pose regression network. Our key contributions are:
\begin{itemize}
\item We study different ways to blend the loss of the student both w.r.t. ground truth and w.r.t. teacher, and propose to use the teacher loss as a confidence score to attentively learn from examples that the teacher is good at predicting through Attentive Imitation Loss (AIL).
\item We also propose Attentive Hint Training (AHT) as a novel way to learn the intermediate representation of the teacher, based on the teacher loss.
\item We perform extensive experiment on KITTI and Malaga datasets which show that our proposed approach can reduce the number of student parameters by up to $92.95\%$ ($2.12\times$ faster) whilst keeping the prediction accuracy very close to that of the teacher.
\end{itemize}

\section{Related Work}

Many compression methods have been recently developed. Besides KD, there are other approaches available such as quantization, pruning, and low-rank decomposition.

Network quantization reduces the number of bits required to compress the network. The quantization could be applied by using 16-bit or 8-bit representation as proposed by \cite{vanhoucke2011improving, gupta2015deep}. As an extreme case, a 1-bit representation (or binary network) could also be used as seen in \cite{Courbariaux2015, Hubara2016a}. By restricting the weights into two possible values (e.g. -1 or 1), binary networks can dramatically reduce both computation time and memory consumption at the cost of significant reduction in accuracy \cite{cheng2018}.

Pruning, as the name implies, removes redundant and non-informative weights. Weight pruning can be done by using magnitude-based method \cite{Han2015, Guo2016} or dropout-based method \cite{Yao2017, Molchanov2017}. The pruning can be applied in the individual neuron connection \cite{Han2015, Han2017} or in convolutional filter itself \cite{luo2017thinet}. This approach promises significant parameter reduction without greatly affecting accuracy, although it typically requires more training stages \cite{Han2017}. However, in practice, it requires additional implementation of sparse matrix multiplication which possibly needs more resource consumption and specialized hardware and software \cite{luo2017thinet}.

Low-rank decomposition reduces DNN complexity by exploiting the low-rank constraints on the network weights. Most approaches, such as \cite{rigamonti2013learning, jaderberg2014speeding, Bhattacharya2016}, factorize a fully connected or convolutional layer through matrix/tensor decomposition techniques such as SVD or CP-decomposition. By using this factorization technique, the number of matrix multiplications becomes smaller than the original network. However, the obtained compression ratio is generally lower than the pruning-based approach and the low-rank constraint imposed on the network might impact the network performance if the rank is not selected with care.

\section{Background: Distillation for Classification}

Knowledege Distillation (KD) is an approach to transfer the knowledge of a large teacher network to a small student network. The main idea of KD is to allow the student to capture the finer structure learned by the teacher instead of learning solely from the true labels. Let $T$ be the teacher network where $\textbf{O}_{T} = \text{softmax} (\textbf{a}_{T}) $ is the teacher output probability and $\textbf{a}_{T}$ is the teacher's logits (pre-softmax output). A student network $S$ with $\textbf{O}_{S} = \text{softmax} (\textbf{a}_{S})$ as the prediction and $\textbf{a}_{S}$ as the logits is trained to mimic $\textbf{O}_{T}$. Since $\textbf{O}_{T}$ is usually very close to the one-hot representation of the class labels, a temperature $\tau > 1$ is used to soften the output probability distribution of $T$. The same temperature is used for training $S$ such that $\textbf{O}_{T}^{\tau} = \text{softmax}(\frac{\textbf{a}_{T}}{\tau})$ and $ \textbf{O}_{S}^{\tau} = \text{softmax}(\frac{\textbf{a}_{S}}{\tau})$, but $\tau = 1$ is then used for testing $S$. If $\mathcal{H}$ is the cross-entropy and $\textbf{y}$ is the one-hot encoding of the true labels, then KD objective function is formed by minimizing both hard label ($\textbf{y}$) error and soft label error (illustrated in Fig. \ref{fig:different_clas_reg} (a)) as follows
\begin{align}
  \label{eq:kd_classification}
  \mathcal{L}_{KD} = \alpha \mathcal{H} ( \textbf{y}, \textbf{O}_{S} ) + (1 - \alpha) \mathcal{H} ( \textbf{O}_{T}, \textbf{O}_{S} ) 
\end{align}
where $\alpha$ is a parameter to balance both cross-entropies.

The KD formulation with softened outputs ($\tau > 1$) in (\ref{eq:kd_classification}) gives more information for $S$ to learn, as it provides information about the relative similarity of the incorrect predictions \cite{Hinton2015}, \cite{Romero2014}. For example, $T$ may mistakenly predict an image of a car as a truck, but that mistake still has a much higher probability than mistaking it for a cat. These relative probabilities of incorrect prediction convey how $T$ tends to generalize to new data \cite{Hinton2015}. However, this advantage does not exist in the regression problem. As seen in Fig. \ref{fig:different_clas_reg} (b), both teacher and ground truth label have the same characteristic. Intuitively, we would prefer to minimize $S$'s prediction directly w.r.t. the ground truth labels since the teacher label is plagued with an unknown error distribution. However, our experiments show that training $S$ only with the ground truth labels gives very poor results. 

\section{Blending Teacher, Student, and Imitation Loss}
\label{sec:blend_teacher_student_imitation_loss}
As it is unclear how we can take advantage of $T$'s prediction in distilling regression networks, we study different ways to blend together the loss of $S$'s prediction w.r.t. ground truth and w.r.t $T$'s prediction. For simplicity, we refer to the error of $S$ w.r.t ground truth as \textit{student loss} and the loss of $T$ w.r.t ground truth as \textit{teacher loss}. We refer to \textit{imitation loss} ($\mathcal{L}_{imit}$) for the errors of $S$ w.r.t. $T$ (since $S$ tries to imitate $T$'s prediction). The following outlines different formulations and rationales of blending teacher, student, and imitation loss.

\noindent \textbf{Minimum of student and imitation}.
In the simplest formulation, we assume that $T$ has good prediction accuracy in all conditions. In this case, as $T$'s prediction will be very close to the ground truth, it does not really matter whether we minimize $S$ w.r.t. ground truth or w.r.t. $T$. Then, we simply minimize whichever one is smaller between the student loss and imitation loss as follows
\begin{align}
  \mathcal{L}_{reg}  & =  \frac{1}{n} \sum_{i=1}^{n} \min \left( \norm{\textbf{p}_{S} - \textbf{p}_{gt}}^2 , \norm{\textbf{p}_{S} - \textbf{p}_{T}}^2 \right) \label{eq:min_reg_loss}
\end{align}
where $\textbf{p}_{S}$, $\textbf{p}_{T}$, and  $\textbf{p}_{gt}$ are $S$'s prediction, $T$'s prediction, and ground truth labels respectively.

\noindent \textbf{Imitation loss as an additional loss}.
Instead of seeking the minimum between the student and imitation loss, we can use the imitation loss as an additional loss term for the student loss. In this case, we regard the imitation loss as another way to regularize the network and prevent the network from overfitting \cite{Hinton2015}. Then, the objective function becomes
\begin{align}
  \mathcal{L}_{reg}  & =  \frac{1}{n} \sum_{i=1}^{n} \alpha \norm{\textbf{p}_{S} - \textbf{p}_{gt}}^2 + (1-\alpha) \norm{\textbf{p}_{S} - \textbf{p}_{T}}^2 \label{eq:add_reg_loss}
\end{align}
where $\alpha$ is a scale factor used to balance the student and imitation loss. This formulation is similar to the original formulation of KD for classification as seen in (\ref{eq:kd_classification}) except the cross-entropy loss is replaced by regression loss.

\noindent \textbf{Teacher loss as an upper bound}.
Equations (\ref{eq:min_reg_loss}) and (\ref{eq:add_reg_loss}) assume that $T$ has very good generalization capability in most conditions. However in practice, $T$ can give very erroneous guidance for $S$. There is a possibility that in adverse environments, $T$ may predict camera poses that are contradictory to the ground truth pose. Hence, instead of directly minimizing $S$ w.r.t. $T$, we can utilize $T$ as an upper bound. This means that $S$'s prediction should be as close as possible to the ground truth pose, but we do not add additional loss for $S$ when its performance surpasses $T$ \cite{chen2017learning}. In this formulation, (\ref{eq:add_reg_loss}) becomes the following equation
\begin{align}
\mathcal{L}_{reg}  & = \frac{1}{n} \sum_{i=1}^{n} \alpha \norm{\textbf{p}_{S} - \textbf{p}_{gt}}^2 + (1-\alpha) \mathcal{L}_{imit} \label{eq:upper_bound_loss} \\
  \mathcal{L}_{imit} & = 
  \begin{cases}
    \norm{\textbf{p}_{S} - \textbf{p}_{T}}^2 , &  \text{if } \norm{\textbf{p}_{S} - \textbf{p}_{gt}}^2 > \norm{\textbf{p}_{T} - \textbf{p}_{gt}}^2 \label{eq:upperbound_condition} \\
    0, & \text{otherwise}
  \end{cases}
\end{align}

\noindent \textbf{Probabilistic imitation loss (PIL)}. 
As stated before, $T$ is not always accurate in practice. Since there is some degree of uncertainty in $T$'s prediction, we can explicitly model this uncertainty with a parametric distribution. For example, we can model the imitation loss using Laplace's distribution
\begin{align}
  {\rm I\!P} \left( \textbf{p}_{S} | \textbf{p}_{T}, \sigma \right)  & = \frac{1}{2 \sigma} \exp \frac{- \norm{\textbf{p}_{S} - \textbf{p}_{T}}}{\sigma}  \label{eq:laplace_dist}
\end{align}
where $\sigma$ is an additional quantity that $S$ should predict. In this case, the imitation loss is turned into minimizing the negative log likelihood of (\ref{eq:laplace_dist}) as follows
\begin{align}
  - \log {\rm I\!P} \left( \textbf{p}_{S} | \textbf{p}_{T}, \sigma \right) & = \frac{ \norm{\textbf{p}_{S} - \textbf{p}_{T}}}{\sigma} + \log \sigma + \text{const.}
  \label{eq:negative_log}
\end{align}
The final objective is retrieved by replacing $\mathcal{L}_{imit}$ in (\ref{eq:upper_bound_loss}) with (\ref{eq:negative_log}). We can view (\ref{eq:negative_log}) as a way for $S$ to learn suitable coefficient (via $\sigma$) to down-weight unreliable $T$'s prediction. Besides Laplacian distribution, another parametric distribution like Gaussian can be used as well.

\noindent \textbf{Attentive imitation loss (AIL)}. 
The main objective of modeling the uncertainty in the imitation loss is that we could then adaptively down-weight the imitation loss when a particular $T$'s prediction is not reliable. However, modeling $T$'s prediction with a parametric distribution may not accurately reflect the error distribution of $T$'s prediction. Hence, instead of relying on $S$ to learn a quantity $\sigma$ to down-weight unreliable $T$'s prediction, we can use the empirical error of $T$'s prediction w.r.t. ground truth (which is the teacher loss) to do the job. Then, the objective function becomes 
\begin{align}
  \mathcal{L}_{reg}  & = \frac{1}{n} \sum_{i=1}^{n} \alpha \norm{\textbf{p}_{S} - \textbf{p}_{gt}}^2_i + (1-\alpha) \Phi_{i} \norm{\textbf{p}_{S} - \textbf{p}_{T}}^2_i \label{eq:attentive_imitation_loss} \\
  \Phi_{i} & = \left( 1 - \frac{ \norm{\textbf{p}_{T} - \textbf{p}_{gt}}^2_i }{\eta} \right) \label{eq:normalized_teacher_loss} \\
  \eta & = \max \left( e_{T} \right)  - \min \left( e_{T} \right) \\
  e_{T} & = \{ \norm{\textbf{p}_{T} - \textbf{p}_{gt}}^2_j : j=1,...,N \} \label{eq:teacher_loss_entire_data}
\end{align}
where $\Phi_i$ is the normalized teacher loss for each $i$ sample, $e_{T}$ is a set of teacher loss from entire training data, and $\eta$ is a normalization parameter that we can retrieve from subtracting the maximum and the minimum of $e_{T}$. Note that $\norm{.}_i$ and $\norm{.}_j$ are not $p$-norm symbol. Instead we use $i$ and $j$ in $\norm{.}_i$ and $\norm{.}_j$ as index to differentiate which loss is computed from the batch samples ($i=1,...,n$) and which loss is calculated from entire training data ($j=1,...,N$).

Fig. \ref{fig:different_clas_reg} (b) shows how each component in (\ref{eq:attentive_imitation_loss})-(\ref{eq:teacher_loss_entire_data}) blends together. Note that we still keep $\alpha$ to govern the relationship between student and imitation loss. In this case, $\Phi_i$'s role is to put different relative importance, hence it is called \textit{attentive}, for each component in the imitation loss as seen in weighted sum operation. Notice that (\ref{eq:attentive_imitation_loss}) can be rearranged into $\mathcal{L}_{reg} = \frac{\alpha}{n} \sum_{i=1}^{n} \norm{\textbf{p}_{S} - \textbf{p}_{gt}}^2_i + \frac{1-\alpha}{n} \sum_{i=1}^{n} \Phi_{i} \norm{\textbf{p}_{S} - \textbf{p}_{T}}^2_i$. As $\Phi_i$ is computed differently for each image sample and is intended to down-weight unreliable $T$'s prediction, we could also say that by multiplying the imitation loss with $\Phi_i$, we rely more on the example data which $T$ is good at predicting in the process of knowledge transfer between $T$ and $S$.

\section{Learning Intermediate Representations}
\label{sec:learn_representation}

Blending teacher, student, and imitation loss have set the objective function for the main KD task. Another important aspect in KD's transfer process is Hint Training (HT). HT is the process of training the intermediate representation of $S$ such that it could mimic the latent representation of $T$. It was designed as an extension of the original KD \cite{Hinton2015} and formulated by \cite{Romero2014} to transfer the knowledge of $T$ to $S$ with deeper but thinner architecture. Even if it is devised to help training $S$ with deeper layers than $T$, we would argue that it is also an important step for training a regressor network with shallow architecture. HT could act as another way to regularize $S$ such that it could better mimic the generalization capability of $T$ \cite{Romero2014}.

In Hint Training, a \textit{hint} is defined as a layer in $T$ that is used to guide a \textit{guided} layer in $S$. Let $\textbf{W}_{guided}$ and $\textbf{W}_{hint}$ be the parameters of $S$ and $T$ up to their guided and hint layers respectively. With the standard HT formulation, we can train $S$ up to the guided layer by minimizing $\mathcal{L}_{hint} = \frac{1}{n} \sum_{i=1}^{n} \norm{\Psi_{T}(\textbf{I};\textbf{W}_{hint}) - \Psi_{S}(\textbf{I};\textbf{W}_{guided})}^2$, where $\Psi_{T}$ and $\Psi_{S}$ are $T$'s and $S$'s deep neural functions up to their respective hint or guided layers. The drawback with this formulation is that it does not take into account the fact that $T$ is not a perfect function estimator and can give incorrect guidance to $S$. While in Section \ref{sec:blend_teacher_student_imitation_loss} we describe how to tackle this issue through down-weighting unreliable $T$ prediction by multiplying it with normalized teacher loss, we argue that this step is also required for HT. Then, we propose a modification of HT termed \textit{Attentive Hint Training} (AHT) as follows:
\begin{align}
 \label{eq:hint_loss_modified}
  \mathcal{L}_{hint} & = \frac{1}{n} \sum_{i=1}^{n} \Phi_i \norm{\Psi_{T}(\textbf{I};\textbf{W}_{hint}) - \Psi_{S}(\textbf{I};\textbf{W}_{guided})}^2_i
\end{align}
where $\Phi_i$ is the normalized teacher loss as seen in (\ref{eq:normalized_teacher_loss}). While (\ref{eq:attentive_imitation_loss}) and (\ref{eq:hint_loss_modified}) can be trained jointly, we found out that training separately yields superior performance especially in absolute pose error. Then, the knowledge transfer between $T$ and $S$ becomes 2 stages optimization procedures. The 1st stage trains $S$ up to the guided layer with (\ref{eq:hint_loss_modified}) as the objective. The 2nd stage trains the remaining layer of $S$ (from guided until the last layer) with (\ref{eq:attentive_imitation_loss}) as the objective.

\section{Implementation Details}

\subsection{Camera Pose Regression with DNNs}
As we demonstrate our distillation approach for Visual Odometry (VO) problem, we will briefly review VO approaches. Conventional VO estimates the camera poses by finding feature correspondences between multiple images and applying multiple-view geometry techniques \cite{Hartley2004, Saputra2018}. On the other hand, DNNs learn the camera ego-motion directly from raw image sequences by training the network in an end-to-end manner. Let $\textbf{I}_{t-1,t} \in {\rm I\!R}^{2 \times ( w \times h \times c)}$ be two concatenated images at times $t-1$ and $t$, where $w$, $h$, and $c$ are the image width, height, and channels respectively. DNNs essentially learn a mapping function to regress the 6-DoF camera poses $\{ ( {\rm I\!R}^{2 \times (w \times h \times c)} )_{1:N} \} \rightarrow \{ ( {\rm I\!R}^{6} )_{1:N} \}$, where $N$ are the total number of image pairs. In the supervised case, learning 6-DoF camera poses can be achieved by minimizing the discrepancy between the predicted poses $\textbf{p}_{pr} \in {\rm I\!R}^{6}$ and the ground truth poses $\textbf{p}_{gt} \in {\rm I\!R}^{6}$ as follows $\mathcal{L}_{reg} = \frac{1}{n} \sum_{i=1}^{n} \norm{\textbf{p}_{pr} - \textbf{p}_{gt}}^2$, given $n$ sample images. However, since translation and rotation have different constraints, we usually decompose $\mathcal{L}_{reg}$ into
\begin{align}
  \mathcal{L}_{reg}  & = \frac{1}{n} \sum_{i=1}^{n} \beta \norm{\textbf{t}_{pr} - \textbf{t}_{gt}}^2 + (1-\beta) \norm{\textbf{r}_{pr} - \textbf{r}_{gt}}^2 \label{eq:trans_rot_loss}
\end{align}
where $\textbf{t} \in {\rm I\!R}^{3}$ and $\textbf{r} \in {\rm I\!R}^{3}$ are the translation and rotation components in $x, y$, and $z$ axes. $\beta \in {\rm I\!R}$ is used to balance $\textbf{t}$ and \textbf{r}. Here the rotation part is represented as an Euler angle. Another representation such as quaternion or rotation matrix can be used as well \cite{wang2017}. 

\begin{figure}[!ht]
    \centering
    \includegraphics[width=4cm,trim=0.2cm .1cm .2cm .7cm,clip]{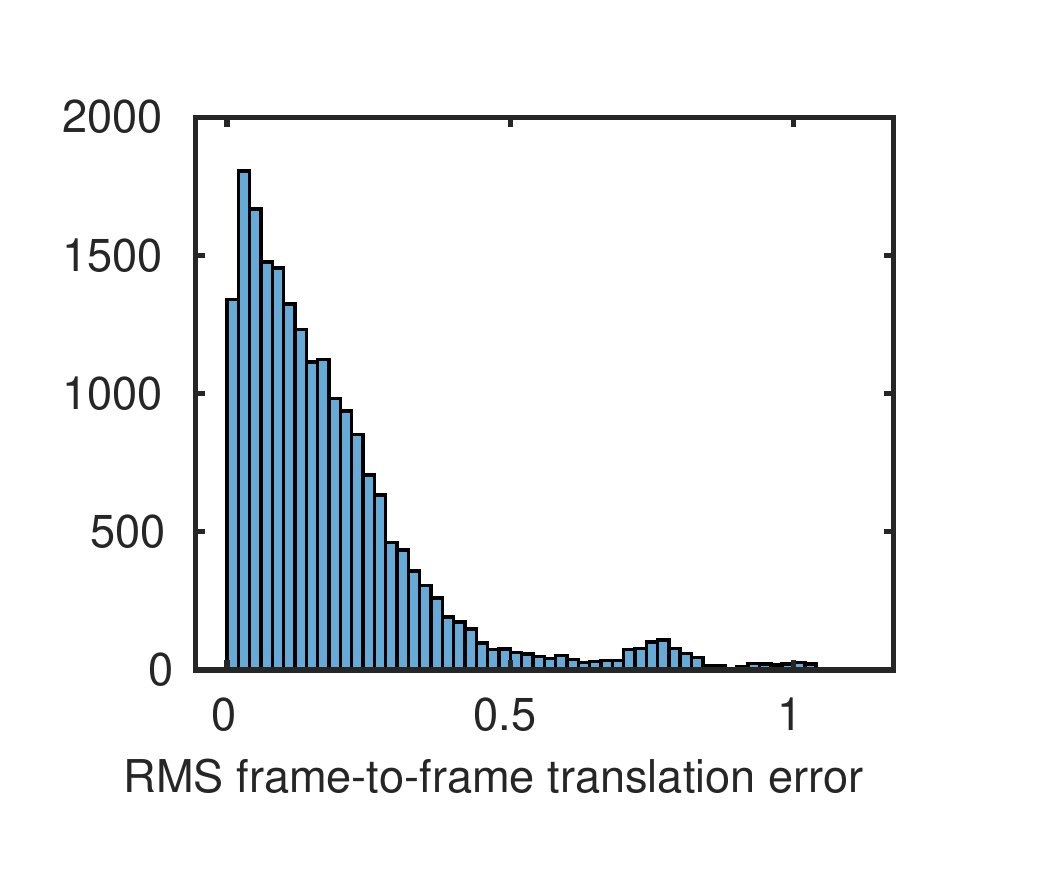}
    \includegraphics[width=4cm,trim=0.2cm .1cm .2cm .7cm,clip]{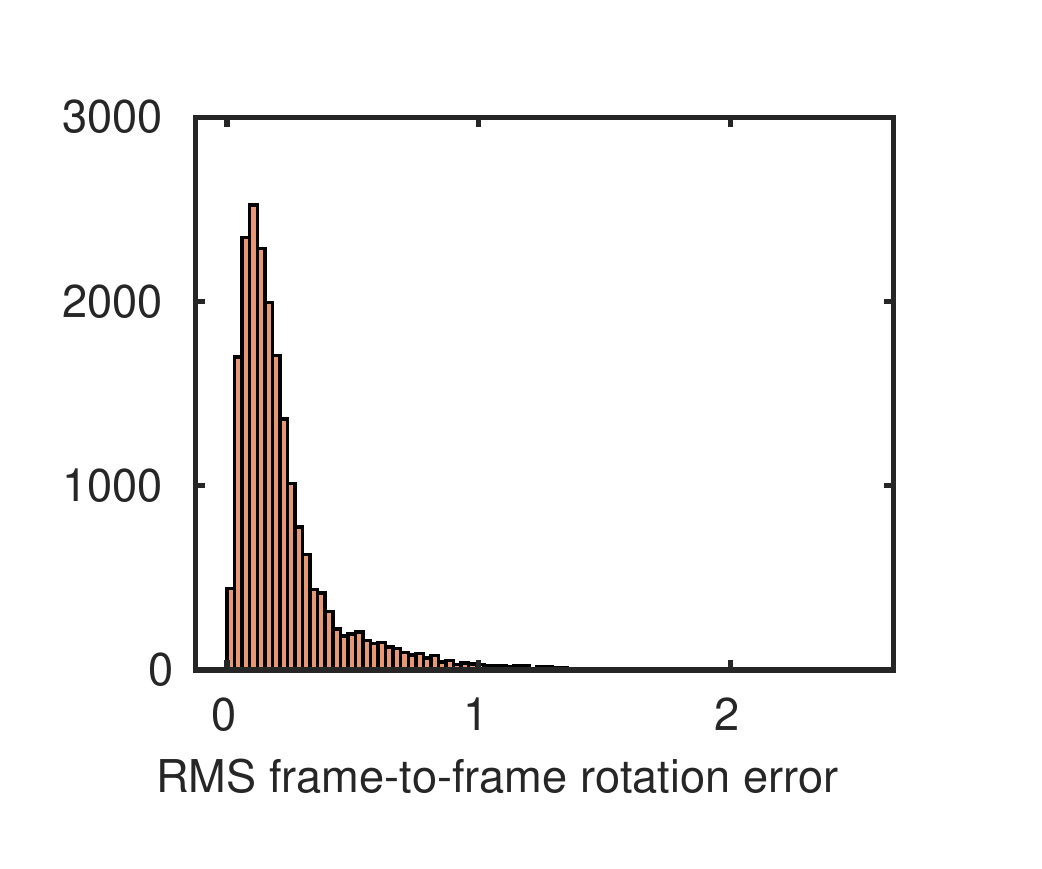}
    \caption{Empirical error distribution of the teacher network for translation and rotation on KITTI dataset Seq 00-08.}
\label{fig:teacher_error_dist}
\end{figure}

To apply our distillation approach to DNN-based VO, we decompose each translation and rotation component in (\ref{eq:trans_rot_loss}) to (\ref{eq:attentive_imitation_loss}). We also decompose AHT formulation in (\ref{eq:hint_loss_modified}) into translation and rotation representation, and apply $\Phi_i$ differently for each representation. As the teacher loss distribution is also different for translation and rotation (as seen in Fig. \ref{fig:teacher_error_dist}), $\eta$ in (\ref{eq:normalized_teacher_loss}) is computed differently for each of them.

\subsection{Network Architecture}
\begin{figure}[!ht]
    \centering
    \includegraphics[width=0.8\columnwidth]{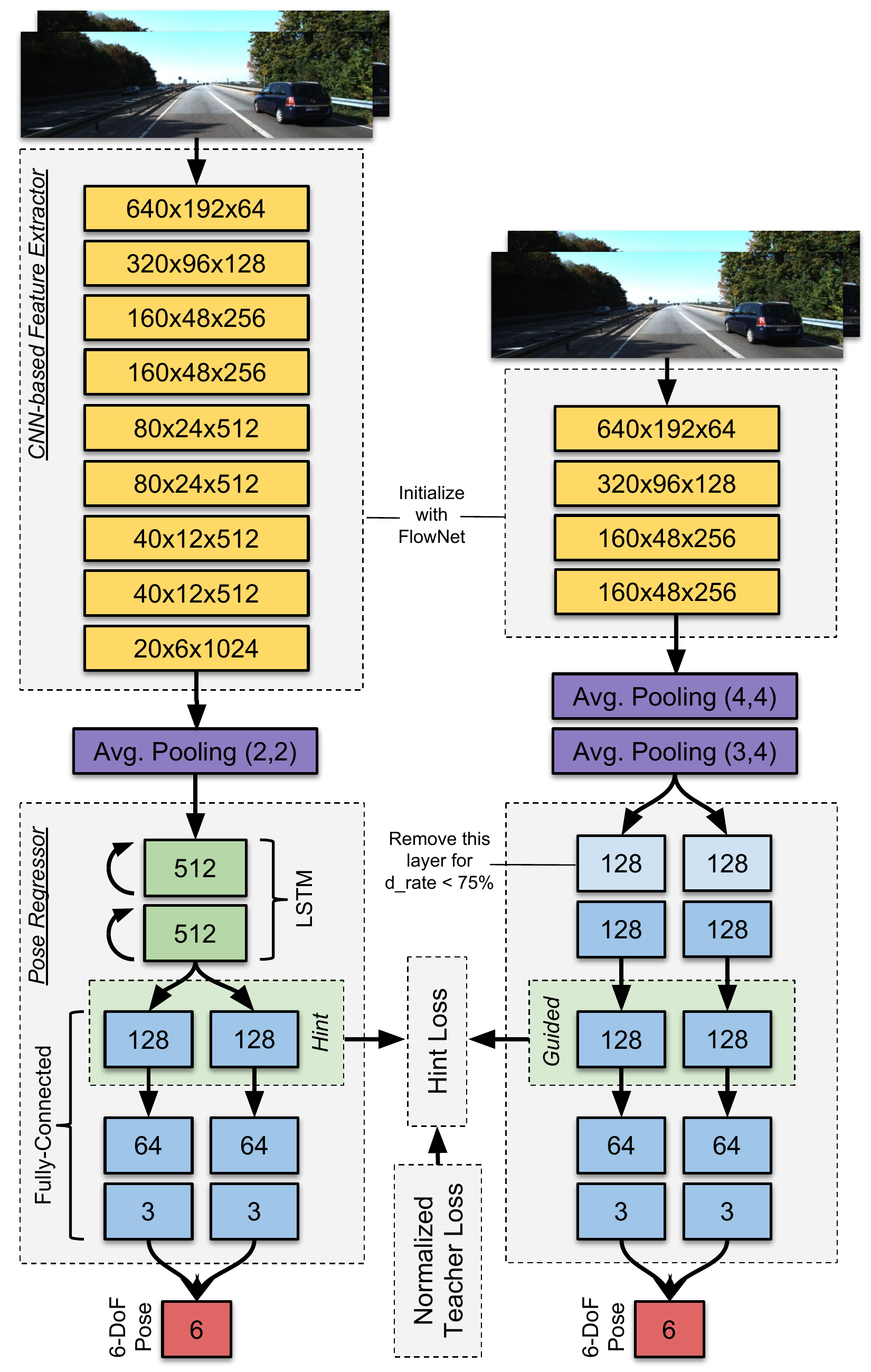}
    \caption{Details of network architecture for teacher (left) and student network with $92.95\%$ distillation rate (right).}
\label{fig:detail_architecture}
\end{figure}

We employ ESP-VO \cite{Wang2017a} for the teacher network $T$ in which the architecture is depicted in Fig. \ref{fig:detail_architecture} (left). It consists of two main parts, namely the feature extractor network and  a pose regressor network. The feature extractor is composed from a series of Convolutional Neural Networks (CNNs) to extract salient features for VO estimation. Since VO estimates the camera pose between consecutive frames, optical-flow like feature extractor network (FlowNet \cite{Dosovitskiy2016}) is used to initialize the CNNs. The pose regressor consists of Long-Short Term Memory (LSTM) Recurrent Neural Networks (RNNs) and Fully Connected (FC) Layers to regress 6-DoF camera poses. The LSTM is utilized to learn long-term motion dependencies among image frames \cite{wang2017}.

Fig. \ref{fig:detail_architecture} (right) depicts $S$ with $92.95\%$ distillation rate ($d_{rate}$). The main building blocks of $S$ are essentially the same as $T$ except we remove a number of layers from $T$ to construct a smaller network. To specify the structure of $S$, in general, we can remove the layers from $T$ which contribute the most to the number of weights, but $S$ should still consist of a feature extractor (CNN) and a regressor (LSTM/FC). In the feature extractor, the largest number of weights usually corresponds to the few last layers of CNNs, while in the regressor part it corresponds to the LSTM layers. Thus, for $d_{rate}=92.95\%$, we remove the last five layers of the CNN and the two RNN-LSTM layers. However, we still initialize the CNN with the FlowNet's weight as in ESP-VO. To compensate for the loss of removing the CNN and LSTM layers, we add 1 FC layer in the regressor part for $d_{rate}<75\%$ and 2 FC layers for $d_{rate}>75\%$. This corresponds to fewer than $1\%$ additional parameters.

\subsection{Training Details}

As stated in Section \ref{sec:learn_representation}, we employ two stages of optimization. The first stage is training the intermediate representation of $S$ through AHT. As seen in Fig. \ref{fig:detail_architecture}, we select the 1st FC layer of $T$ as a hint and the 3rd FC layer of $S$ (or the 2nd FC layer for $d_{rate}<75\%$) as the guided layer. We used the FC layer of $T$ as a hint not only to provide easier guidance for training $S$, since both FC layers in $T$ and $S$ have the same dimensions, but also to transfer the ability of $T$ to learn the long-term motion dynamics of camera poses as the FC layer of $T$ is positioned after the RNN-LSTM layers. In the second stage, we freeze weights of $S$ trained from the first stage and train the remaining layers of $S$ using (\ref{eq:attentive_imitation_loss}) as the objective.

\section{Experimental Results}

\subsection{Experiment Environments}
We implemented $T$ and $S$ in Keras. We employed NVIDIA TITAN V GPU for training and NVIDIA Jetson TX2 for testing. The training for each stage goes up to 30 epochs. For both training stages, we utilize Adam Optimizer with $1e-4$ learning rate. We also applied Dropout \cite{srivastava2014dropout} with $0.25$ dropout rate for regularizing the network. For the data, we used KITTI \cite{Geiger2012a} and Malaga odometry dataset \cite{blanco2014}. We utilized KITTI Seq 00-08 for training and Seq 09-10 for testing. Before training, we reduced the KITTI image dimension to $192 \times 640$. We only use Malaga dataset for testing the model that has been trained on KITTI. For this purpose, we cropped the Malaga images to the KITTI image size. Since there is no ground truth in Malaga dataset, we perform qualitative evaluation against GPS data.

\subsection{Metrics}
In this work, we want to measure the trade-off between accuracy and parameter reduction. In VO, accuracy can be measured by several metrics. We use Root Mean Square (RMS) Relative Pose Error (RPE) for translation (\textbf{t}) and rotation (\textbf{r}) and RMS Absolute Trajectory Error (ATE) as they has been widely used in many VO or SLAM benchmarks \cite{sturm2012benchmark}. For parameter reduction, we measure the percentage ($\%$) of $S$'s parameters w.r.t. $T$'s parameters. We also measure the associated computation time (ms) and the model size (MB) for each reduction rate.

\subsection{Sensitivity Analysis}
\noindent \textbf{The Impact of Different Methods for Blending Teacher, Student, and Imitation Loss}. In this experiment, we want to understand the impact of different approaches to blending teacher, student, and imitation loss as described in Section \ref{sec:blend_teacher_student_imitation_loss}. We used $S$ with $d_{rate} = 72.78\%$ constructed from removing the last 3 CNNs and replacing 2 LSTMs with 1 FC layer. In order to get a fair comparison without having bias from the AHT process, we trained $S$ with standard HT approach in the first stage. Then, we trained the remaining layer(s) with all formulations described in Section \ref{sec:blend_teacher_student_imitation_loss} in the second stage. For additional comparison, we add a baseline approach, in which we only minimize the student loss.
\begin{figure}[!ht]
    \centering
    \subfloat[Frame-to-frame relative RMSE]{
        \begin{tabular}[b]{c}%
        \includegraphics[width=7.8cm, trim=0.1cm .5cm .1cm .7cm,clip]{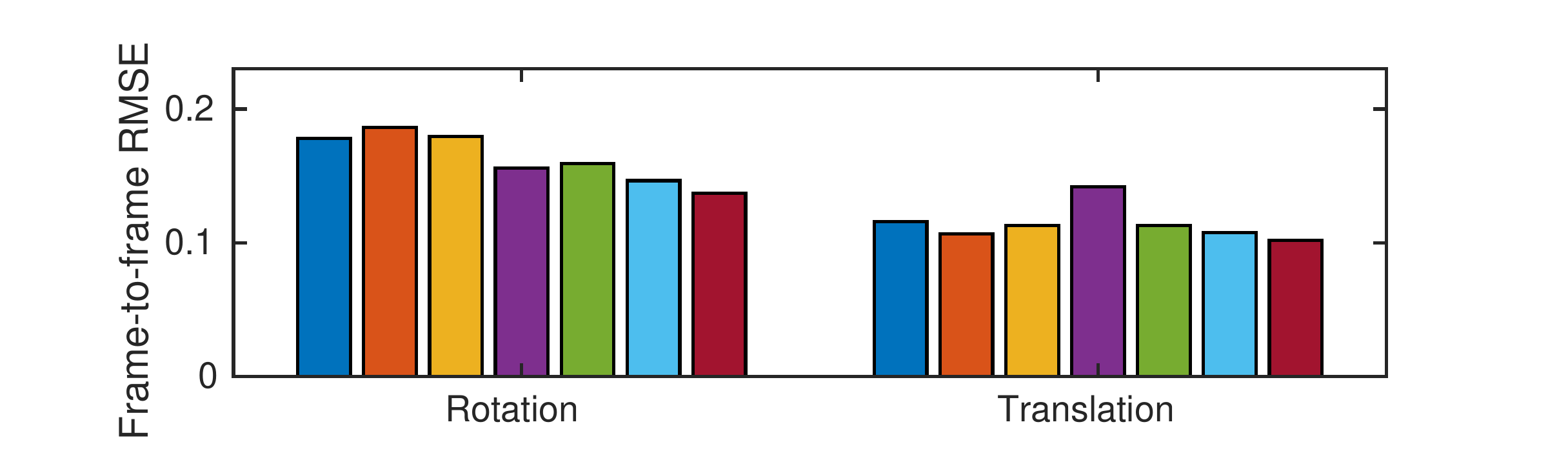}
        \end{tabular}
        }
        \\
    \subfloat[CDF of Absolute Trajectory Errors]{
        \begin{tabular}[b]{c}%
    	\includegraphics[width=7.8cm, trim=0.1cm .7cm .1cm 1.4cm,clip]{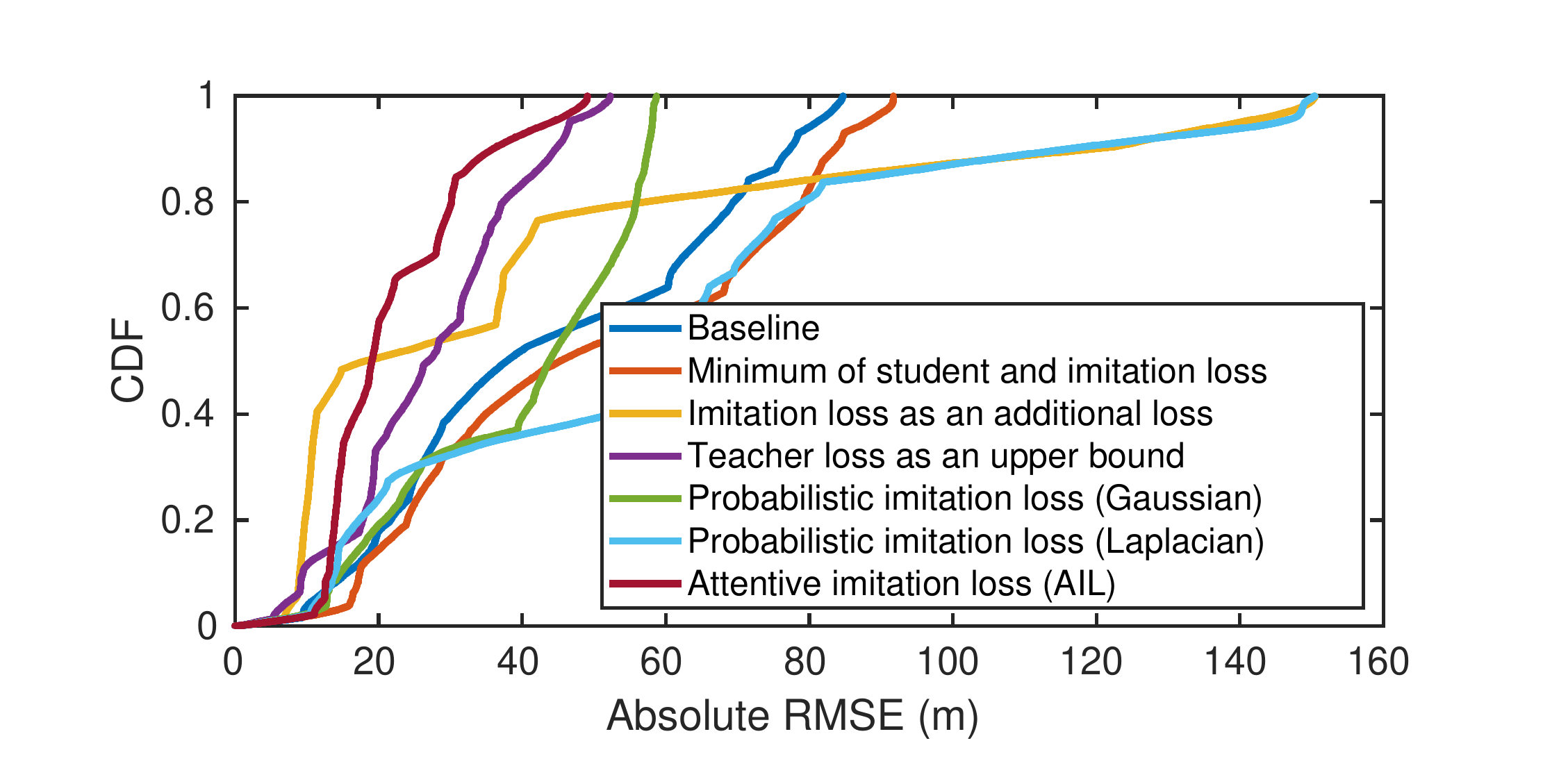}
        \end{tabular}
        }
    \caption{The impact of different ways in blending teacher, student, and imitation loss to (a) RPE and (b) ATE. Same legend is used for both graphs.}
\label{fig:impact_blending_hard_soft}
\end{figure}

Fig. \ref{fig:impact_blending_hard_soft} (a) and (b) depicts the RPE and the CDF of ATE of different methods in blending the losses. It can be seen that AIL has the best accuracy in both RPE and ATE. This indicates that distilling knowledge from $T$ to $S$ only when we trust $T$ does not reduce the quality of knowledge transfer, but instead improve the generalization capability of $S$. Two approaches (minimum of student and imitation; imitation loss as additional loss) that rely on the assumption that $T$'s prediction is always accurate have inferior performance even if compared to the baseline. PIL, either using Laplacian or Gaussian, yields good accuracy in RPE, but lacks robustness since they have larger overall drift (as seen in Fig. \ref{fig:impact_blending_hard_soft} (b)). This is probably due to the failure of the parametric distribution function to model the teacher error distribution accurately. The upper bound objective has good balance between RPE and ATE but the performance is inferior to AIL.
\begin{table}[t]
\centering
\begin{threeparttable}
	\caption{The impact of using Attentive Imitation Loss (AIL) and Attentive Hint Training (AHT) algorithm}
	\renewcommand{\arraystretch}{1}
	\label{table:impact_aht}
	\fontsize{9}{12}\selectfont
	\setlength\extrarowheight{-2pt}
	\begin{tabular}{c|ccccc}
		\hline
		\noalign{\smallskip}
		& Architecture & HT  & Final & Rec. & ATE\\
		& CNN-FC\textsuperscript{\textbf{a}} &  & Obj. & Error\textsuperscript{\textbf{b}} & (m) \\
		\hline
		\noalign{\smallskip}
		1 & 6 - 2 (72.88\%) & - & student & 0.6242 & 80.842 \\
		2 & 6 - 2 (72.88\%) & HT & student & 0.0252 & 65.251 \\
		3 & 6 - 2 (72.88\%) & HT & \underline{AIL} & 0.0252 & 36.459 \\
		4 & 6 - 2 (72.88\%) & \underline{AHT} & student & \textbf{0.0166} & 52.320 \\
		5 & 6 - 2 (72.88\%) & \underline{AHT} & \underline{AIL} & \textbf{0.0166} & \textbf{32.259} \\
		\hline
		\noalign{\smallskip}
		6 & 5 - 3 (79.69\%) & - & student & 0.1341 & 68.350 \\
		7 & 5 - 3 (79.69\%) & HT & student & 0.0177 & 53.661 \\
		8 & 5 - 3 (79.69\%) & HT & \underline{AIL} & 0.0177 & 29.751 \\
		9 & 5 - 3 (79.69\%) & \underline{AHT} & student & \textbf{0.0168} & 37.645 \\
		10 & 5 - 3 (79.69\%) & \underline{AHT} & \underline{AIL} & \textbf{0.0168} & \textbf{25.857} \\
		\hline
	\end{tabular}
	\begin{tablenotes}
    \fontsize{7}{9}\selectfont
    \item[\textbf{a}] Total FC layers until intermediate layer used for HT and AHT. The number in the bracket indicates $d_{rate}$.
    \item[\textbf{b}] Reconstruction error of $S$'s output intermediate representation w.r.t. $T$'s output intermediate representation.
    \end{tablenotes}
\end{threeparttable}
\end{table}

\noindent \textbf{The Impact of Attentive Hint Training}. As we want to inspect the effect of the proposed AHT approach, we trained the model with 3 different procedures: without HT, with HT, and with AHT. We also alternate between using the student loss and AIL to see the effect of applying attentive transfer mechanism in both intermediate (as AHT) and final layer (as AIL), or only in one of them. We used the same model architecture as the previous ablation to conduct this experiment. We compare RMS Reconstruction Error of $S$'s output latent representation w.r.t. $T$'s representation and ATE w.r.t. ground truth.

\begin{figure}[!ht]
    \centering
    \includegraphics[width=9.2cm, trim=3.3cm .1cm .1cm .1cm,clip]{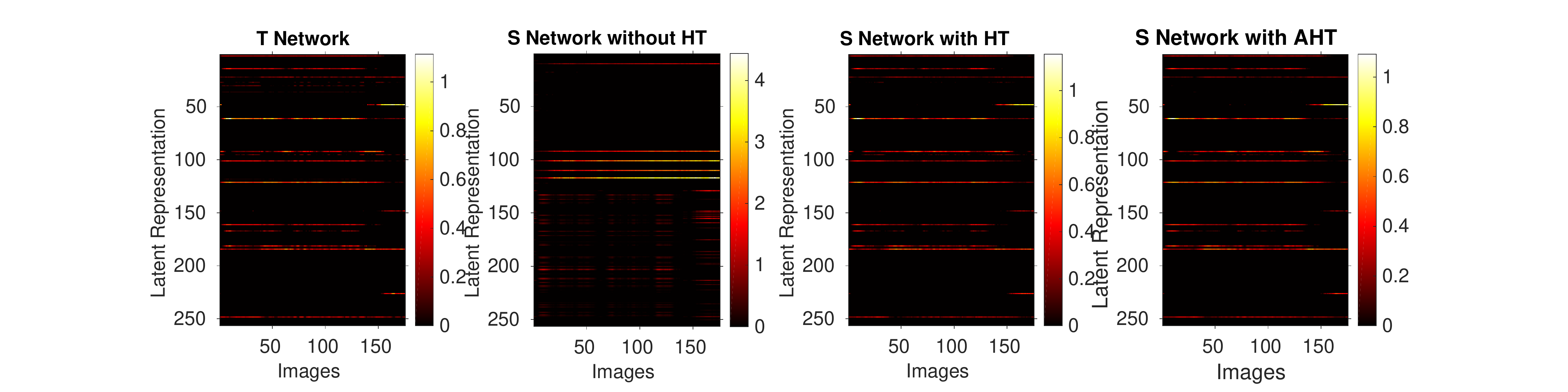}
    \caption{The difference of latent feature representation between $T$ and $S$, trained without HT, with HT, and with AHT.}
\label{fig:vis_feature_representation}
\end{figure}

Table \ref{table:impact_aht} lists the results of this study which clearly shows that as soon as HT is applied, $S$'s reconstruction error w.r.t. $T$ reduces dramatically (see row [1, 2] or [6, 7]). This shows that without having guidance in the intermediate layer, it is very difficult for $S$ to imitate the generalization capability of $T$. AHT then reduces further the reconstruction error of HT by giving different relative importance to $T$'s representation and placing more emphasis on representations that produce accurate $T$ predictions. Fig. \ref{fig:vis_feature_representation} visualizes the output latent representation for different training procedures. It can be seen that AHT's output representation is very close to $T$. Slight differences with $T$'s representation are due to different relative importance placed on $T$'s predictions. However, even if AHT does not try to blindly imitate $T$'s representation, the average reconstruction error is still lower than the HT approach which attempts to perfectly imitate $T$'s representation (see Table \ref{table:impact_aht} row [2, 4] or [7, 9]).


The last column of Table \ref{table:impact_aht} shows ATE for different combinations of applying attentive knowledge transfer. As it can be seen in row [2, 4] (or [7, 9]) that applying attentive loss in the intermediate layer (AHT) can significantly reduce the ATE of $S$. However, the reduction rate is not as large as when applying it in the final task (AIL) (see Table \ref{table:impact_aht} row [2, 3] or [7, 8]) as it can reduce the ATE up to $1.8\times$ smaller. This is sensible because the accuracy of a DNN model depends on the output from the final task. A better guidance in the final layer (main task) can yield stronger performance than a better guidance in the intermediate layer. Finally, applying attentive loss in both intermediate (AHT) and final layers (AIL) consistently gives the best result for 6 and 5 CNNs architecture (see Table \ref{table:impact_aht} row 5 and 10).

\begin{figure}[!ht]
    \centering
    \includegraphics[width=7cm, trim=0.cm .5cm .1cm .9cm,clip]{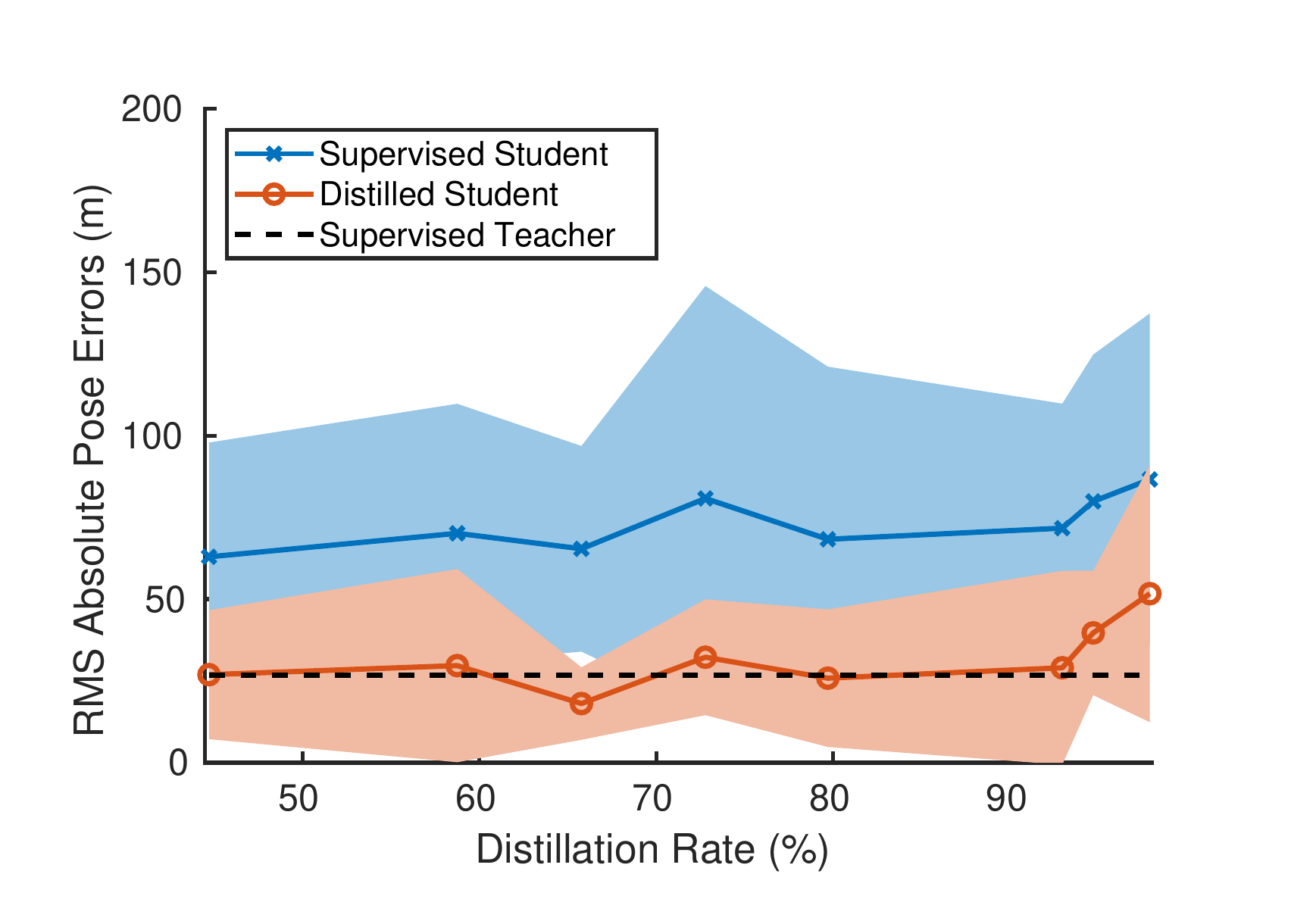}
    \caption{RMS absolute pose errors between Supervised and Distilled Student for different $d_{rate}$.}
\label{fig:tradeoff_compress_acc}
\end{figure}

\begin{figure}[!t]
    \centering
    \includegraphics[width=0.32\columnwidth]{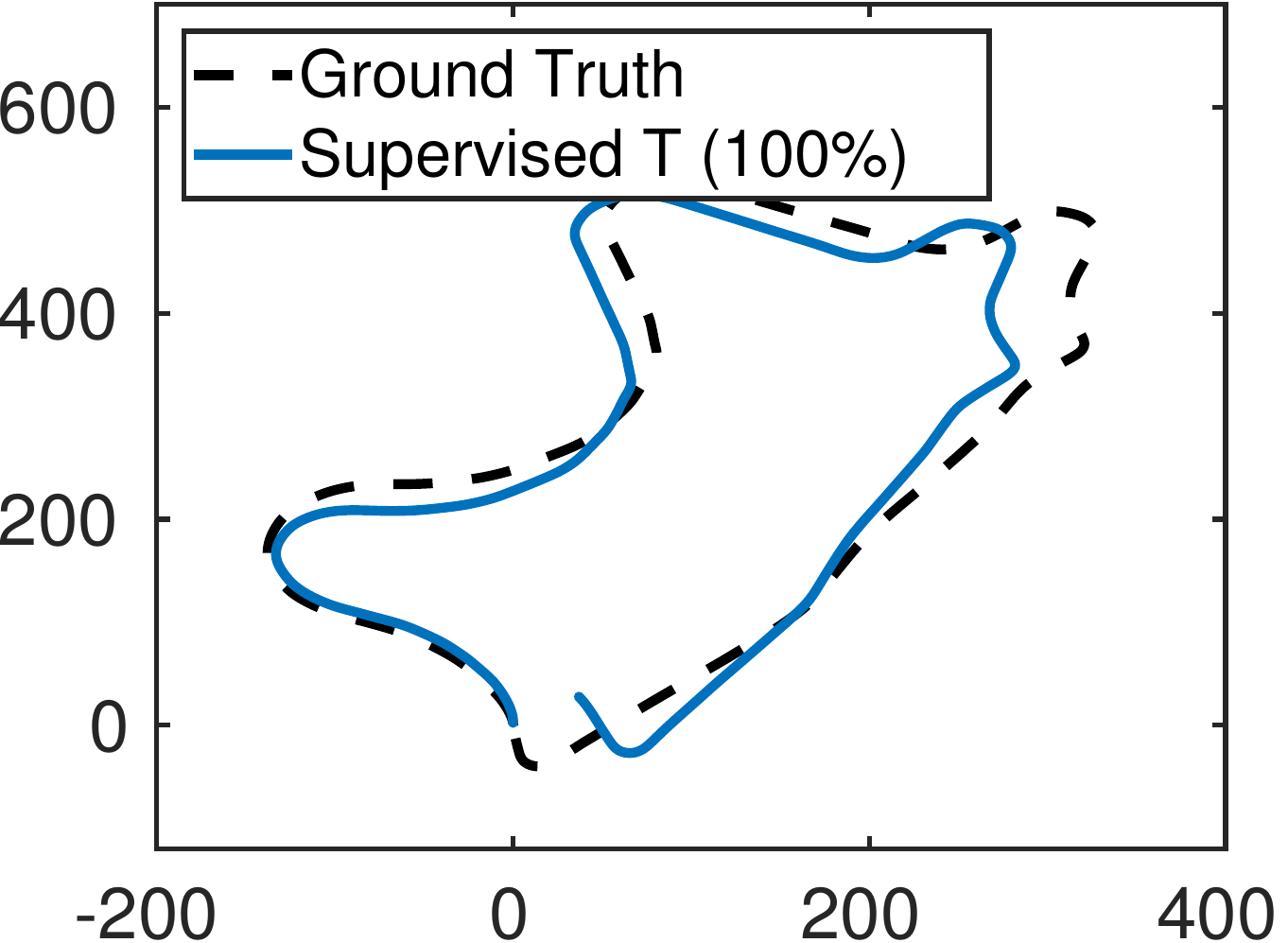}
    \includegraphics[width=0.31\columnwidth]{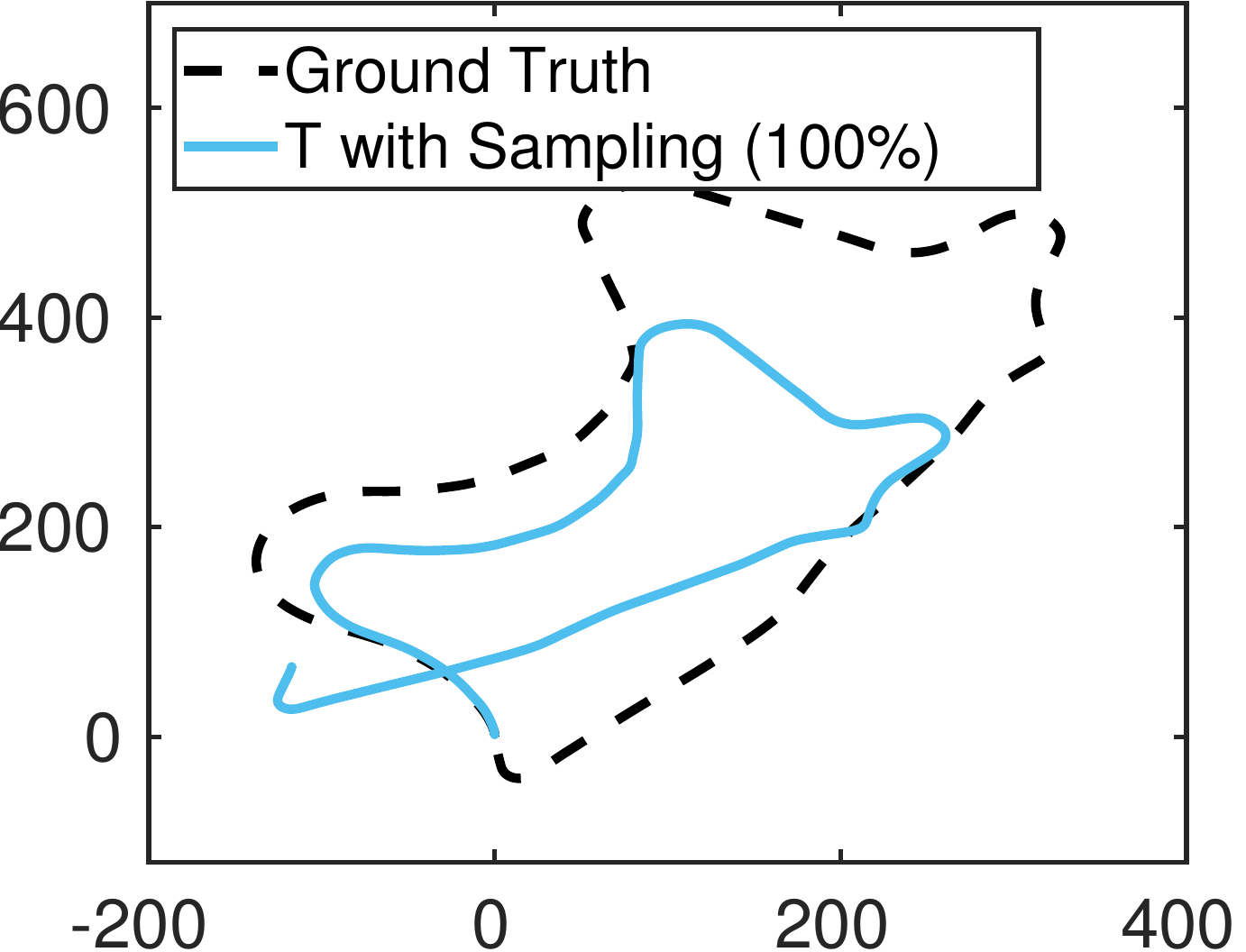}
    \includegraphics[width=0.31\columnwidth]{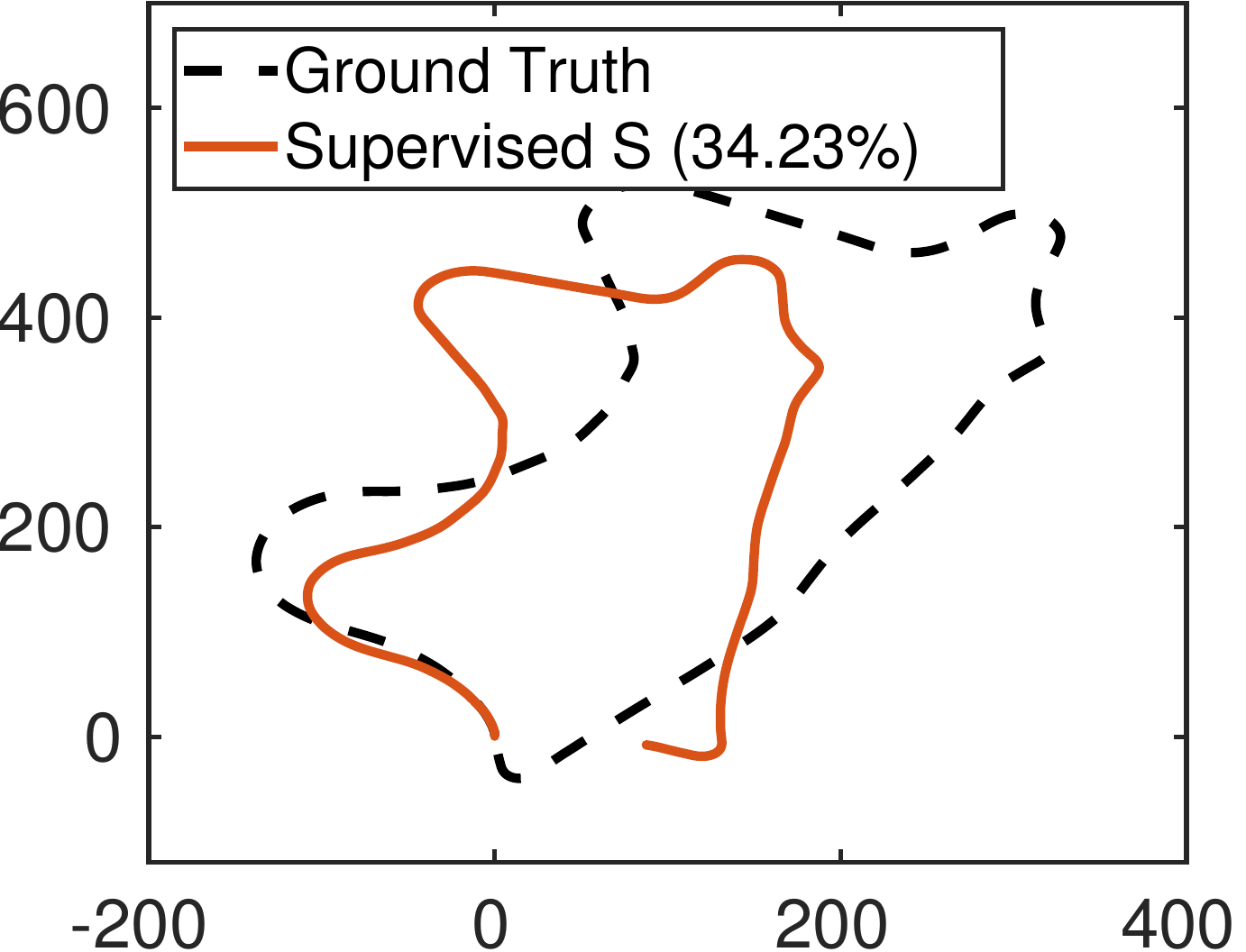} \\
    \vspace{0.2cm}
    \includegraphics[width=0.31\columnwidth]{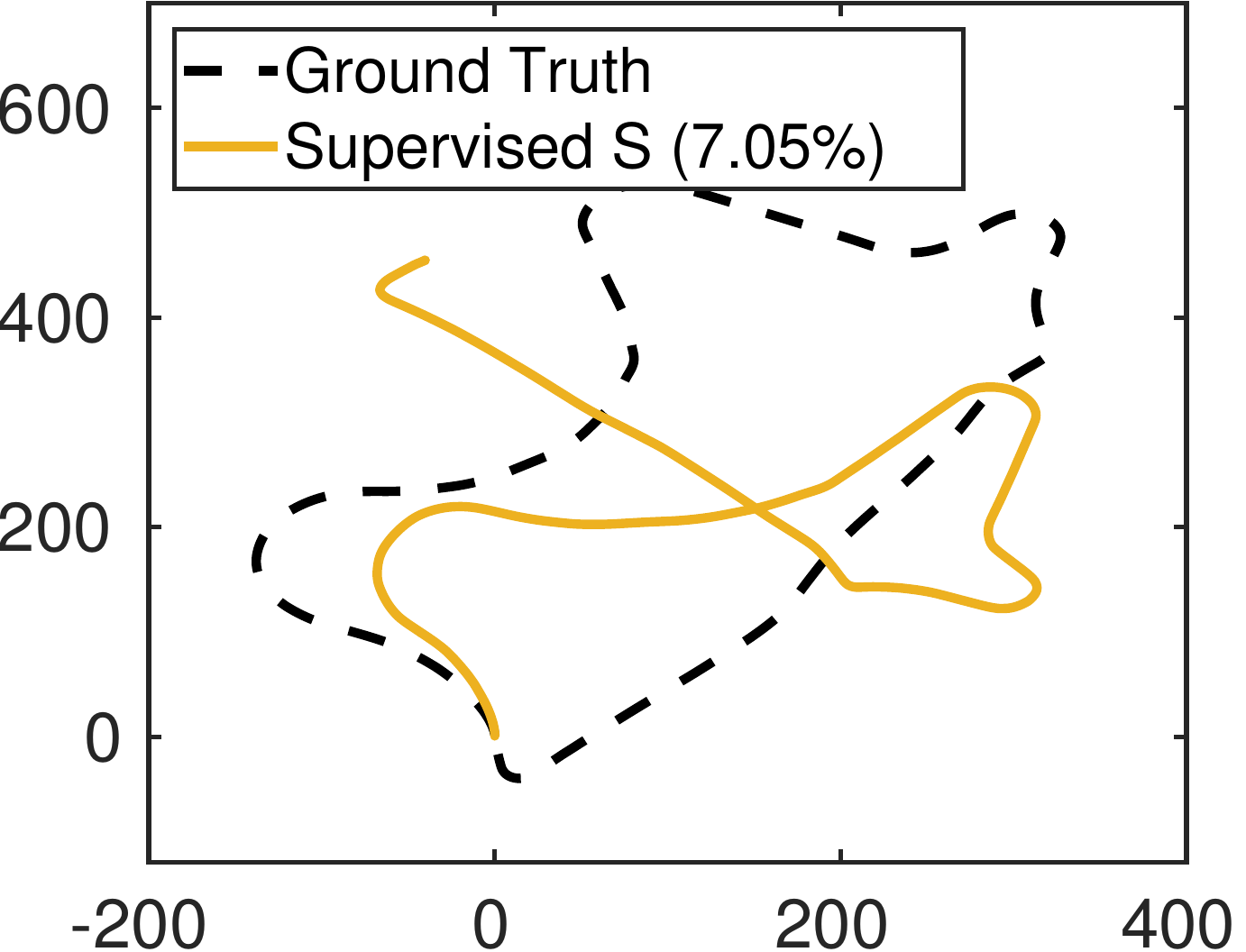}
    \includegraphics[width=0.32\columnwidth]{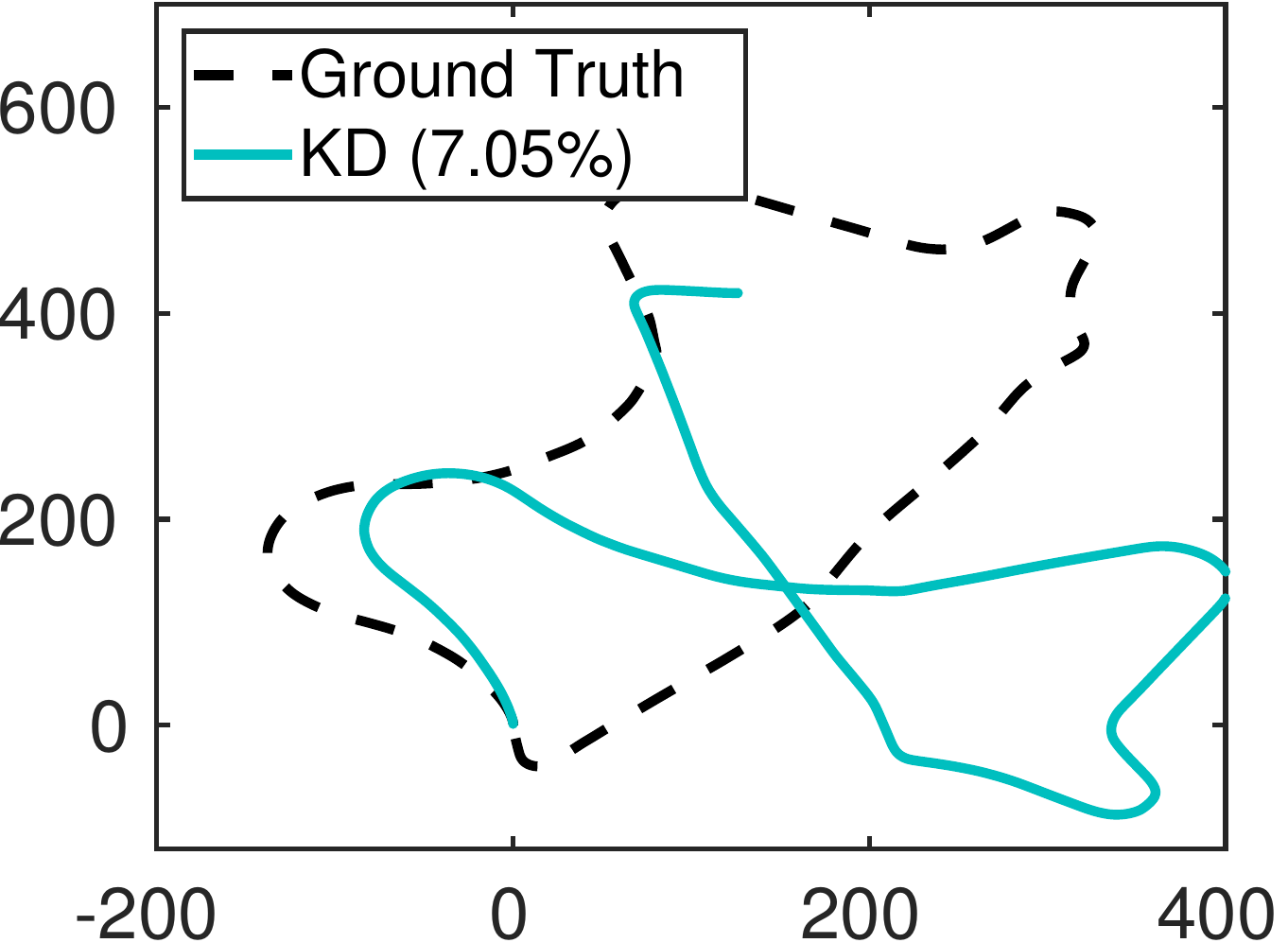}
    \includegraphics[width=0.31\columnwidth]{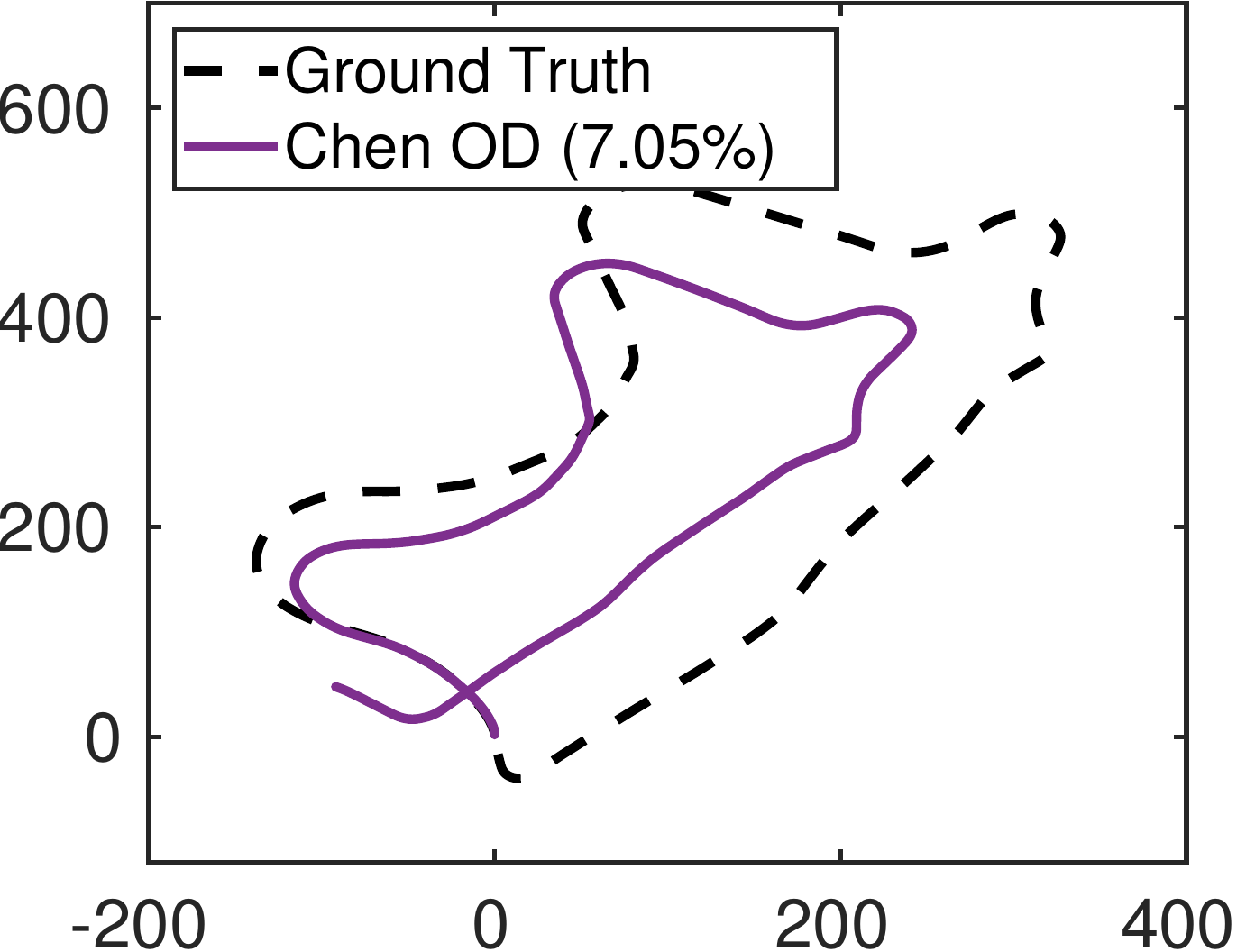}\\
    \vspace{0.2cm}
    \includegraphics[width=0.31\columnwidth]{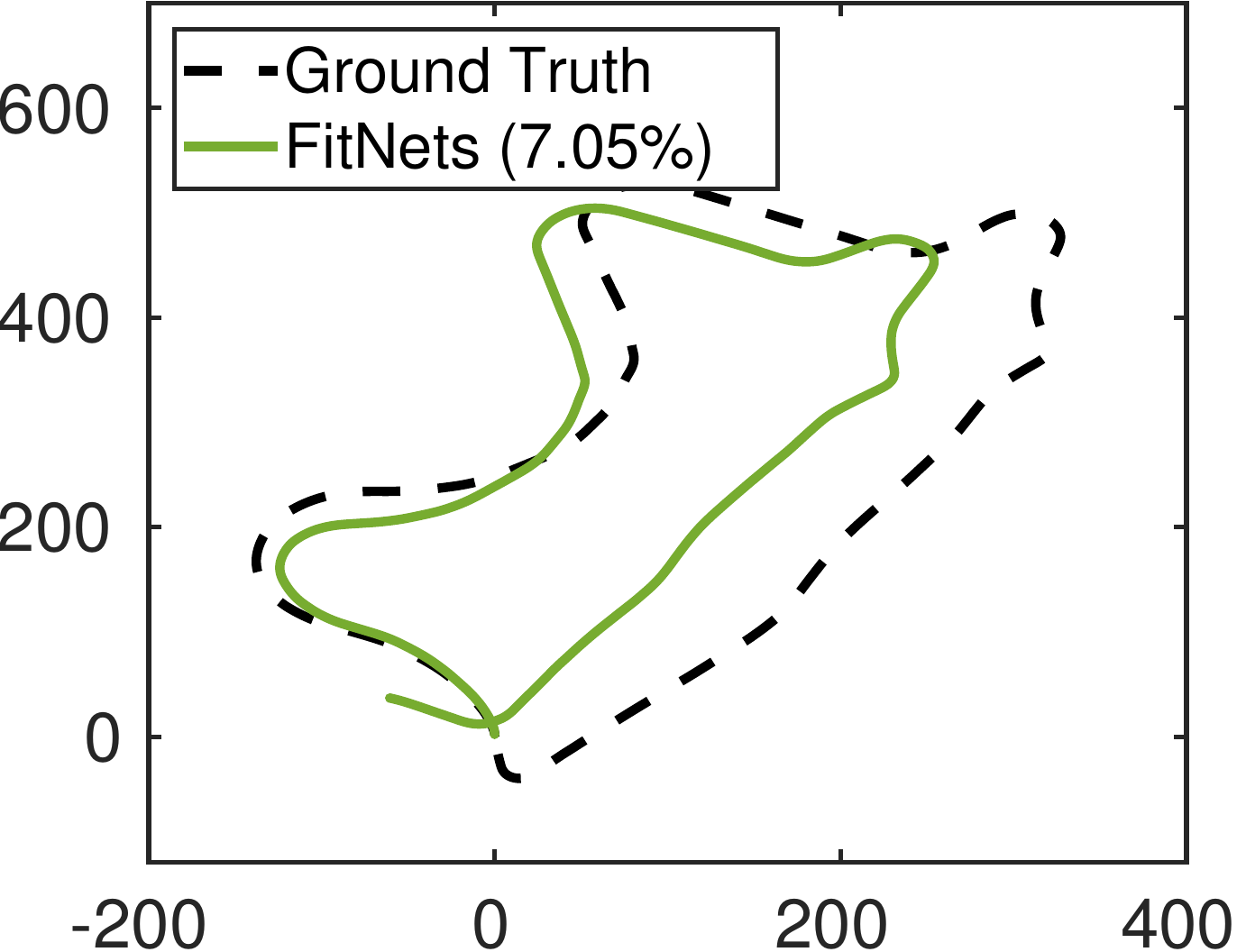}
    \includegraphics[width=0.32\columnwidth]{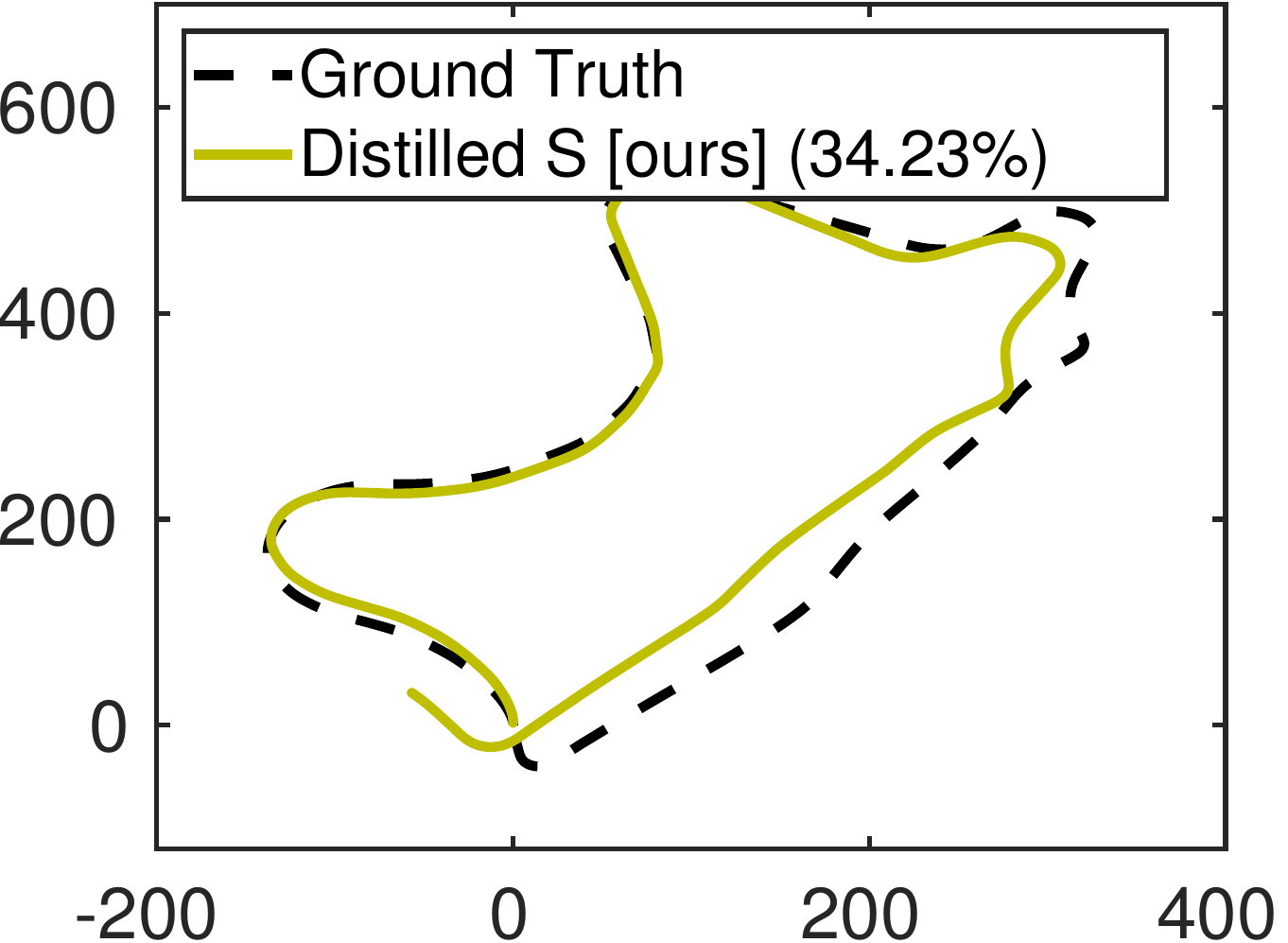}
    \includegraphics[width=0.31\columnwidth]{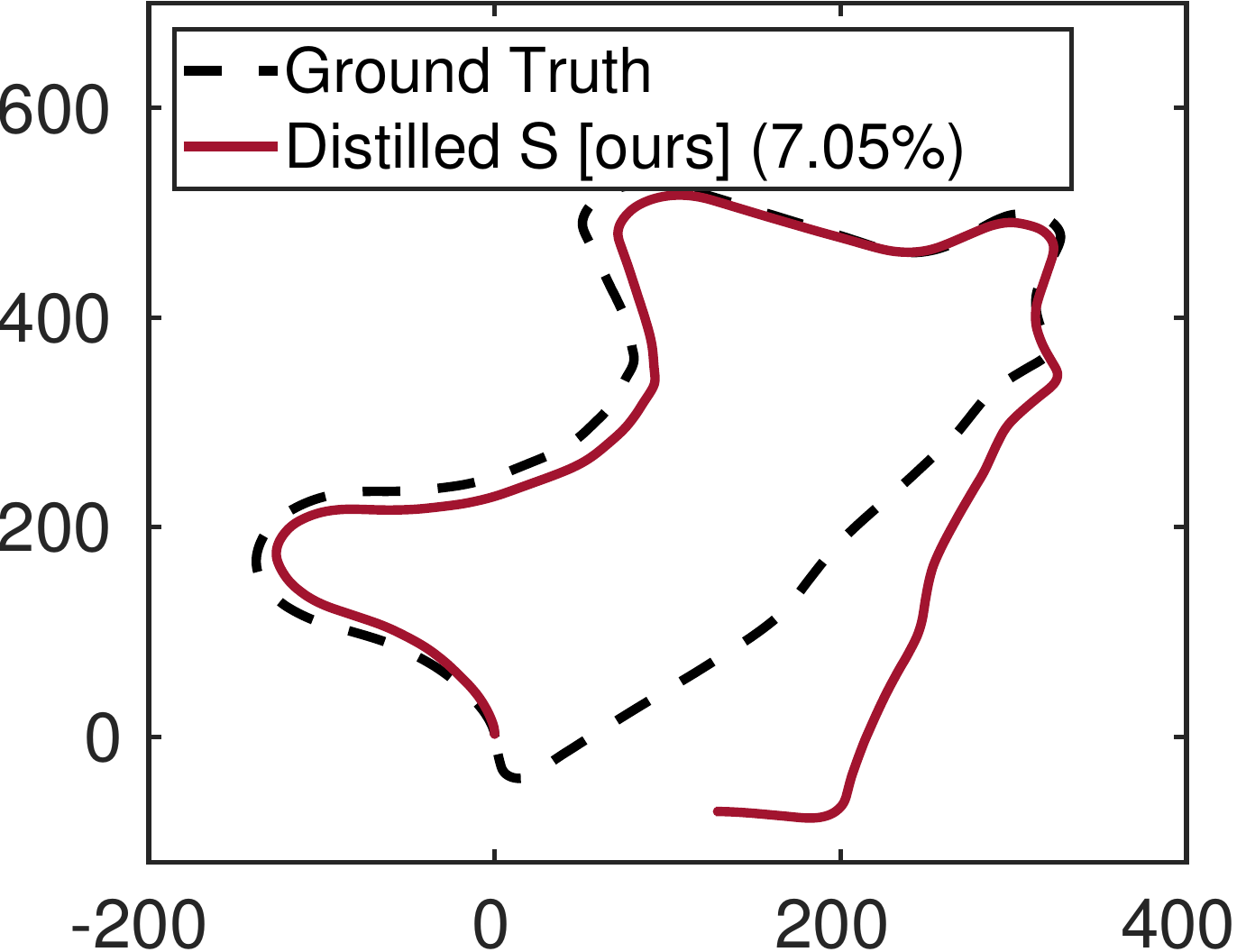}
    \caption{Trajectory prediction from $T$ and $S$ trained with various distillation approaches in KITTI Seq 09. The number in the bracket indicates the percentage of $S$ parameters w.r.t. $T$.}
\label{fig:trajectory}
\end{figure}


\subsection{Trade-off between Accuracy, Model Size, and Execution Time}
\label{sec:tradeoff_acc_compress}
In this experiment, we want to understand the trade-off between the model size, execution time, and accuracy for different $d_{rate}$. Fig. \ref{fig:tradeoff_compress_acc} shows that our proposed distillation approach can keep $S$ very close to $T$ up to $d_{rate} = 92.95\%$. It can even achieve better performance than the teacher for $d_{rate} = 65.77\%$ and $79.69\%$ as $T$ might be over-parameterized (see also the output trajectory in Fig. \ref{fig:trajectory}). For $d_{rate} > 92.95\%$, the performance starts to degrade more rapidly as it becomes too difficult to transfer the knowledge from $T$ to $S$ without other constraints. It can also be seen that if $S$ is trained directly to fit the ground truth with hard loss (supervised student), it shows very poor performance.
\begin{table}[t]
\centering
\begin{threeparttable}
	\caption{Trade-off between the number of parameters, model size, computation time, and accuracy (ATE)}
	\renewcommand{\arraystretch}{1}
	\label{table:param_vs_time}
	\fontsize{9}{12}\selectfont
	\setlength\extrarowheight{-3pt}
	\begin{tabular}{ccccc}
		\hline
		\noalign{\smallskip}
		Network & Parameters & Size & Ex. Time & ATE \\
		(Weights $\%$) & (millions) & (MB) & (ms) & (m) \\
		\hline
		\noalign{\smallskip}
		$T$ (100\%) & 33.64 & 286.9 & 87 & 26.74 \\
		$S$ (55.28\%) & 18.59 & 74.6 & 82 & 26.92 \\
		$S$ (41.25\%) & 13.88 & 55.7& 71 & 29.69 \\
		$S$ (34.23\%) & 11.52 & 46.3 & 62 & 18.09 \\
		$S$ (27.22\%) & 9.16 & 36.8 & 58 & 32.26 \\
		$S$ (20.30\%) & 6.83 & 27.5 & 47 & 25.86 \\
		$S$ (7.05\%) & 2.37 & 7.3 & 41 & 29.03 \\
		\hline
	\end{tabular}
\end{threeparttable}
\end{table}

\begin{figure}[!ht]
    \centering
    \includegraphics[width=4cm,trim=0.7cm .3cm .7cm .9cm,clip]{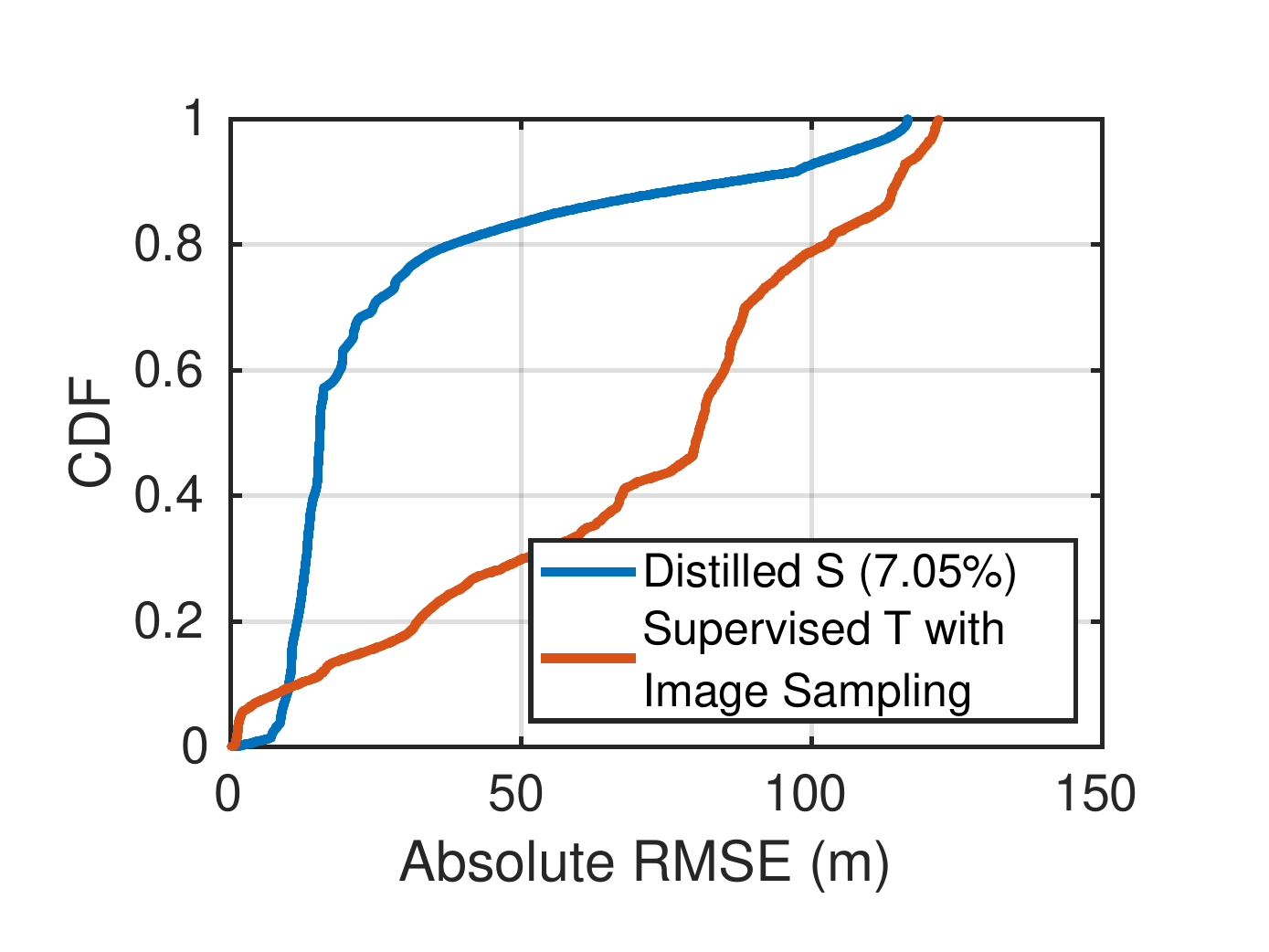}
    \includegraphics[width=4.cm,trim=0.7cm .3cm .7cm .9cm,clip]{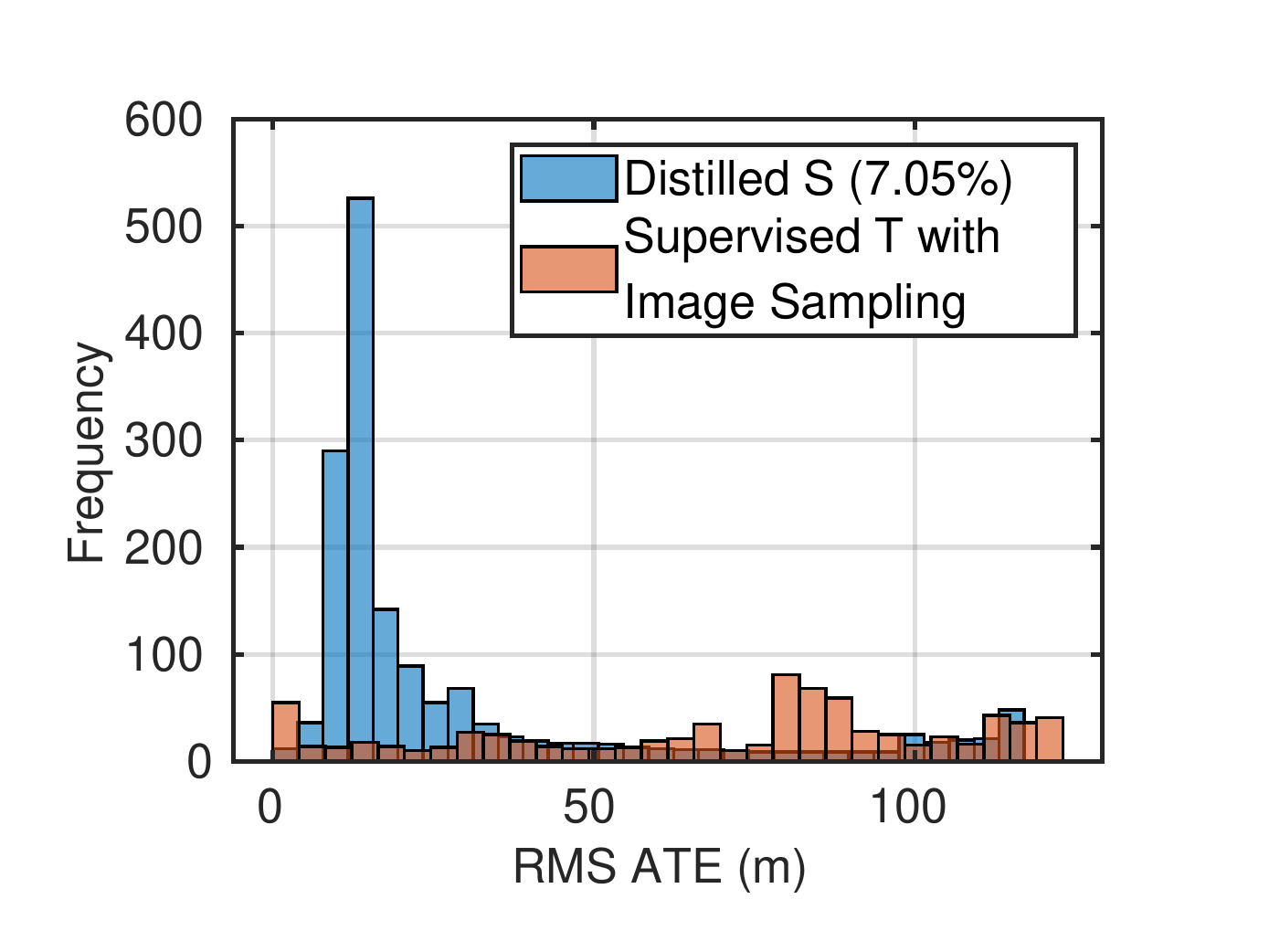}
    \caption{Distribution and histogram of ATE between distilled $S$ and supervised $T$ with image sampling.}
\label{fig:hist_teacher_sampling}
\end{figure}

Table \ref{table:param_vs_time} shows the comparison between $T$ and $S$ in terms of number of parameters, model size, and computation time. As we can see, with $d_{rate} = 92.95\%$ we can reduce the model size from 286.9MB to 7.3MB ($2.5\%$). Removing 2 LSTMs, which are responsible for $44.72\%$ of $T$'s parameters can already reduce $T$'s model size to 78MB ($27\%$) but it has less impact in the computation time as the LSTM has been implemented efficiently for NVIDIA cuDNN. With $d_{rate} = 92.95\%$, we reduce the computation time from 87ms to 41ms ($2.12\times$), effectively doubling the frame rate. This has significant practical implication. If we re-train $T$ given subsampled images such that the frame rate is similar to $S$ with $d_{rate} = 92.95\%$, $T$'s prediction accuracy will degrade $160\%$ (see the trajectory in Fig. \ref{fig:trajectory}). This is probably due to the difficulty of estimating accurate optical flow representation in large stereo baseline. Meanwhile with the same computation budget, the distilled $S$ yields stronger performance than the subsampled $T$ with only $8.63\%$ accuracy drop w.r.t supervised $T$ as seen in Fig. \ref{fig:hist_teacher_sampling}.

\begin{table}[t]
\centering
\begin{threeparttable}
	\caption{Comparison with other distillation approaches}
	\renewcommand{\arraystretch}{1}
	\label{table:comparison_other_distillation}
	\fontsize{9}{12}\selectfont
	\setlength\extrarowheight{-3pt}
	\begin{tabular}{lccc}
		\hline
		\noalign{\smallskip}
		Method & RMS RPE ($\textbf{t}$) & RMS RPE ($\textbf{r}$) & RMS ATE \\
		\hline
		\noalign{\smallskip}
		Supervised $T$ & 0.1197 & 0.2377 & 26.7386 \\
		Supervised $S$ & 0.1367 & 0.1627 & 71.7517 \\
		\hline
		\noalign{\smallskip}
		KD \cite{Hinton2015} & 0.1875 & 0.1439 & 165.2182 \\
		Chen's OD \cite{chen2017learning} & 0.1197 & 0.1416 & 46.2320 \\
		FitNets \cite{Romero2014} & 0.1450 & 0.1409 & 31.9624 \\
		Ours & \textbf{0.1053} & \textbf{0.1406} & \textbf{29.0294} \\
		\hline
	\end{tabular}
\end{threeparttable}
\end{table}

\subsection{Comparison with Related Works}
As there is no specific KD for pose regression, to compare our proposed KD with other related works, we used some well known KD approaches for classification and object detection. However, we modify the objective function to fit our regression problem. We used 3 baselines for this experiment: KD \cite{Hinton2015}, FitNets \cite{Romero2014}, and Chen's Object Detection (OD) model \cite{chen2017learning}. For KD \cite{Hinton2015}, we trained with standard training procedure (without HT or AHT) and replaced the objective function with (\ref{eq:add_reg_loss}). For FitNets \cite{Romero2014}, we used HT approach for the 1st stage of training and utilize (\ref{eq:add_reg_loss}) as the objective in the 2nd stage. For Chen's OD \cite{chen2017learning}, we also used standard HT for the 1st stage and employ (\ref{eq:upper_bound_loss}) as the objective in the 2nd stage. For all models, we used $d_{rate} = 92.95\%$ to train $S$ as an extreme example (see supplementary material for experiment with $d_{rate} = 65.77\%$).

Table \ref{table:comparison_other_distillation} shows the result of this experiment. It can be seen that our proposed approach have better accuracy for both RPE and ATE. Even if most of the competing approaches have better RPE than the supervised $T$, it has huge bias in the relative pose prediction such that the integration of these relative poses yields very large ATE. $T$ tackle this bias to some extent by using LSTM layers which are supposed to learn the long-term motion dynamic of the camera poses \cite{wang2017}. Since $S$ removes the LSTM layers, most approaches fail to recover this knowledge from $T$ but our proposed approach is able to reduce ATE by focusing to learn the good predictions from $T$. Fig. \ref{fig:trajectory} shows the comparison of the output trajectory between the competing approaches on KITTI dataset. It can be seen that our distillation output trajectory is closer to $T$ than the competing approaches.

\begin{figure}[!t]
    \centering
    \includegraphics[width=4.cm,trim=0.cm 0.7cm 6.cm 16.6cm,clip]{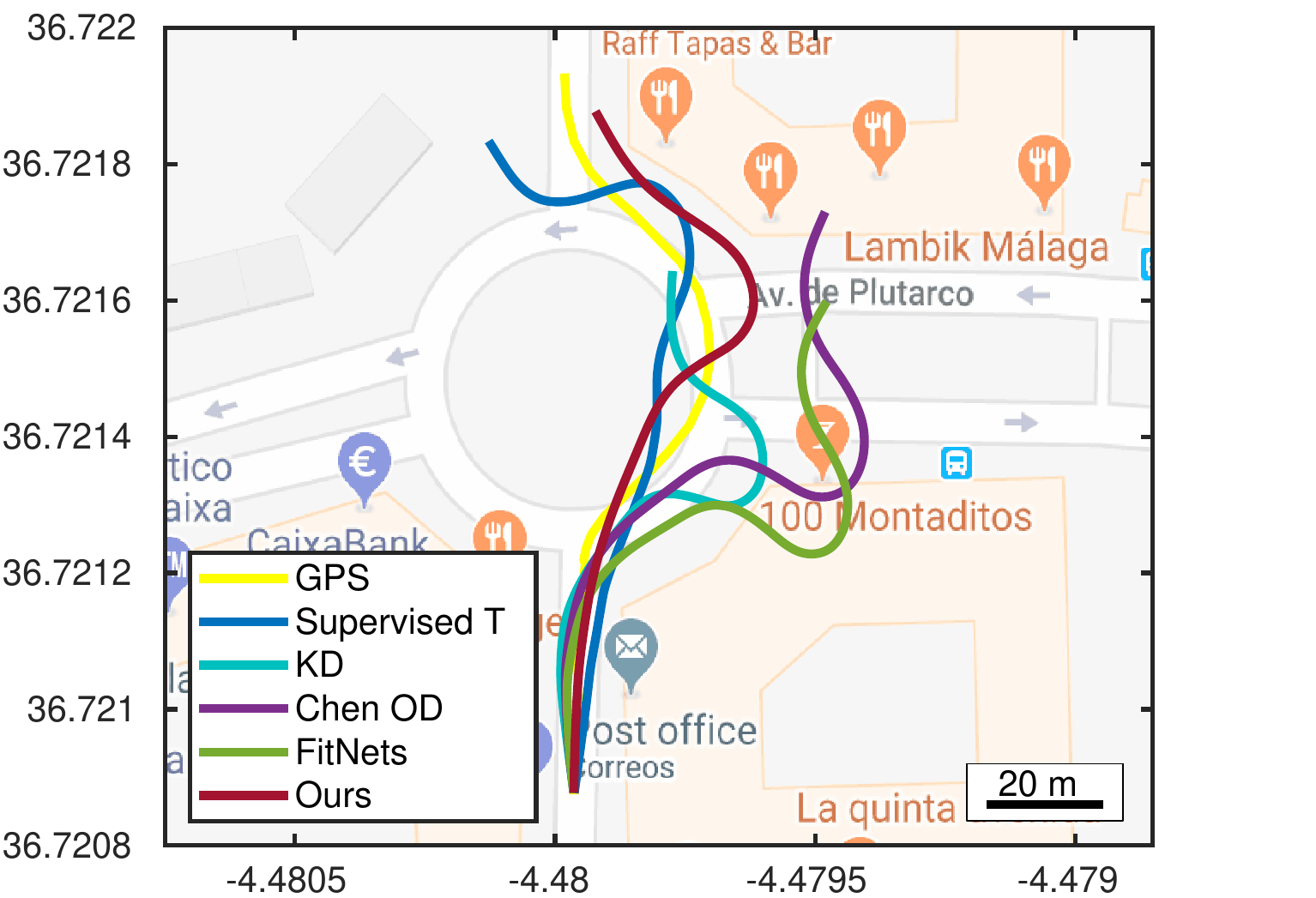}
    \includegraphics[width=4.cm,trim=0.cm 0.7cm 6.cm 16.6cm,clip]{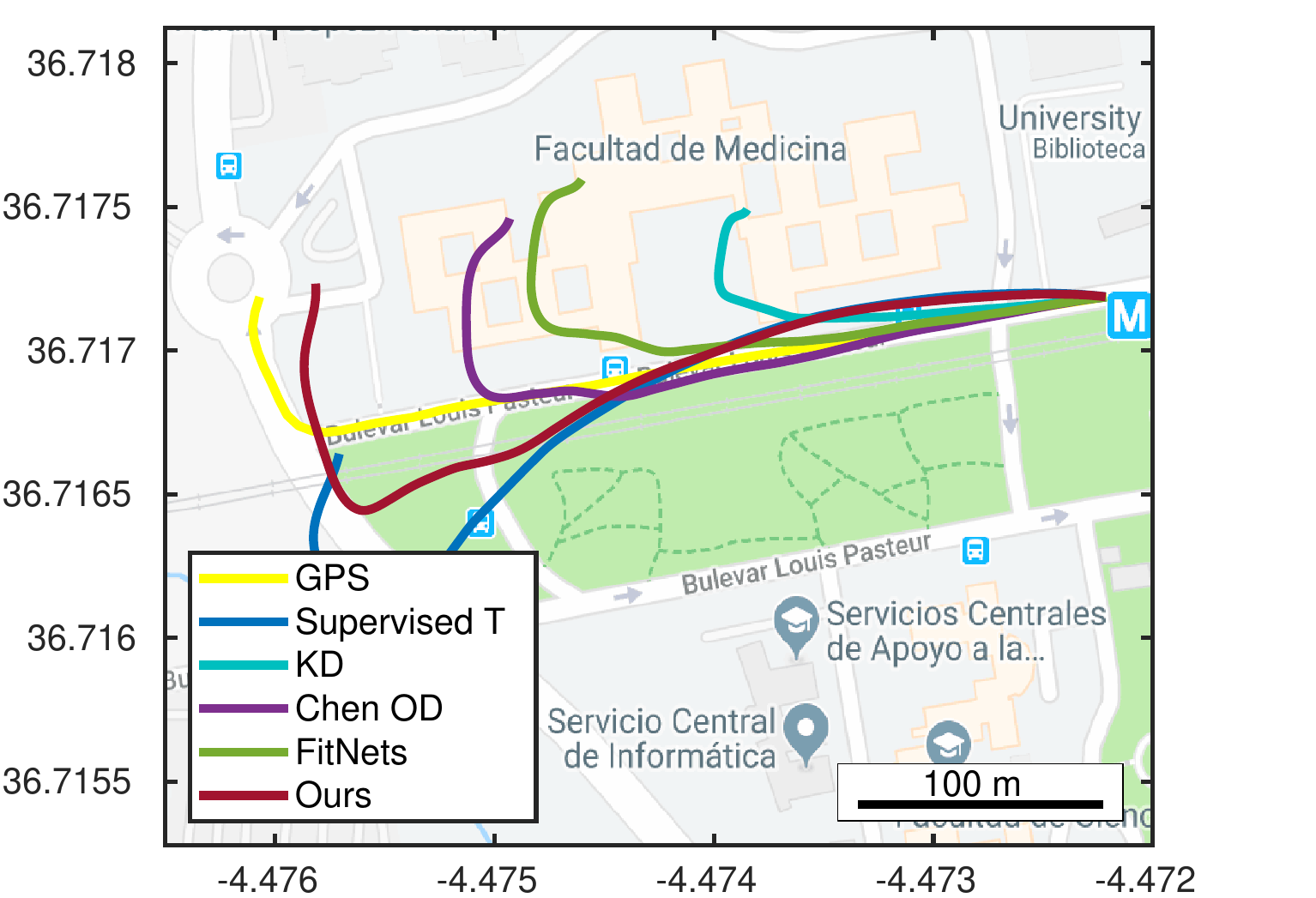}
    \caption{Qualitative evaluation in Malaga dataset Seq 04 and Seq 09. All model are only trained on KITTI Seq 00-08.}
\label{fig:trajectory_malaga}
\end{figure}

Fig. \ref{fig:trajectory_malaga} depicts the qualitative evaluation in Malaga dataset. Note that we have not trained on this dataset, demonstrating generalization capacity. We can see that our proposed approach yields a closer trajectory to GPS, even when it is compared to $T$. This signifies that our distillation approach yields good generalization ability even when it is tested on a different dataset. This result also shows that $T$ may overfit the training data as it has many redundant parameters. However, this redundancy seems necessary for the initial stage of training as a DNN requires large degree-of-freedom to find better weights and connections \cite{Hinton2015}. Meanwhile, directly training $S$ without any supervision from $T$ seems to be very difficult. Our results show that we can have better generalization when distilling large degree-of-freedom $T$ to small degree-of-freedom $S$ if we transfer the knowledge from $T$ to $S$ only when we trust $T$.


\section{Conclusion}
We have presented an approach to distill the knowledge from a deep pose regressor network to a much smaller network with a small loss of accuracy. We have shown that the teacher loss can be used as an effective attentive mechanism to transfer the knowledge between teacher and student. For future work, we will investigate whether another compression technique can be combined with our distillation approach to further reduce the computation time. \\
\textbf{Acknowledgement}. This research is funded by the US
National Institute of Standards and Technology (NIST) Grant
No. 70NANB17H185. M. R. U. Saputra was supported by Indonesia Endowment Fund for Education (LPDP).

{\small
\bibliographystyle{ieee}
\bibliography{library}
}

\end{document}